%% file: paper.tex
\documentclass{article} 

\PassOptionsToPackage{numbers, compress}{natbib}
\input{math_commands.tex}

\usepackage[preprint]{neurips_2019}
\usepackage{hyperref}
\usepackage{cleveref}
\usepackage{url}
\usepackage{bm}
\usepackage{booktabs}
\usepackage{graphicx}
\usepackage{subfigure}
\usepackage{algpseudocode}
\usepackage{booktabs}  
\usepackage{wrapfig,lipsum}
\usepackage{threeparttable} 
\usepackage{multirow}
\usepackage{amsthm}
\usepackage{algorithmicx,algorithm}
\usepackage[superscript]{cite}
\newtheorem{definition}{Definition}
\newtheorem{theorem}{Theorem}
\newtheorem{corollary}{Corollary}
\newtheorem{lemma}{Lemma}
\title{Higher-order Weighted Graph Convolutional Networks}

\author{Songtao Liu\textsuperscript{1,2}, Lingwei Chen\textsuperscript{2}, Hanze Dong\textsuperscript{3}, Zihao Wang\textsuperscript{2}, Dinghao Wu\textsuperscript{2}, Zengfeng Huang\textsuperscript{1,}\thanks{ Corresponding author} \\
\textsuperscript{1}School of Data Science, Fudan University\\
\textsuperscript{2}College of Information Sciences and Technology, The Pennsylvania State University\\
\textsuperscript{3}Departments of Mathematics, The Hong Kong University of Science and Technology\\
\texttt{\{stliu15,huangzf\}@fudan.edu.cn}\\
\texttt{\{lvc5613,zzw166,duw12\}@psu.edu}\\
\texttt{hdongaj@ust.hk}
}

\begin{document}

\maketitle

\begin{abstract}

Graph Convolution Network (GCN) has been recognized as one of the most effective graph models for semi-supervised learning, but it extracts merely the first-order or few-order neighborhood information through information propagation, which suffers performance drop-off for deeper structure. Existing approaches that deal with the higher-order neighbors tend to take advantage of adjacency matrix power. In this paper, we assume a seemly trivial condition that the higher-order neighborhood information may be similar to that of the first-order neighbors. Accordingly, we present an unsupervised approach to describe such similarities and learn the weight matrices of higher-order neighbors automatically through Lasso that minimizes the feature loss between the first-order and higher-order neighbors, based on which we formulate the new convolutional filter for GCN to learn the better node representations. Our model, called higher-order weighted GCN (HWGCN), has achieved the state-of-the-art results on a number of node classification tasks over Cora, Citeseer and Pubmed datasets.
\end{abstract}

\section{Introduction}
Convolutional neural networks (CNNs) have made great achievements on a wide range of image tasks including image classification \citep{simonyan2015very, szegedy2015going, he2016deep, huang2017densely}, object detection \citep{girshick2014rich, redmon2016you, liu2016ssd, dai2016r, lin2017feature}, semantic segmentation \citep{long2015fully, badrinarayanan2017segnet, chen2017deeplab}, etc. Due to the fact that their underlying data representation has a grid-like structure, CNNs perform highly effective on image processing, and can thus capture local patterns by compressing the hypothesis space and using local filters to learn the parameters. However, lots of real-word data cannot be represented as grid-like structure. For example, social networks and biological networks are usually represented as graphs instead of grid-like structure, while the data defined on 3D meshes is important for many graphical applications. As a result, there is an increasing number of fields that focus on studying non-Euclidean structured data.

To address such challenge, inspired by the great success of applying CNNs to computer vision tasks, many research efforts have been devoted to a paradigm shift in graph learning that generalizes convolutions to the graph domain. More specifically, the graph structure is encoded using a convolutional neural network model to operate the neighborhood of each node in graphs. In general, attempts in this direction can be categorized into non-spectral (spatial) approaches and spectral approaches. While recent works are making progress on these two lines of research respectively, here, we focus on the extension of the graph convolution spectral filter. In this respect, the base work that applies
a localized neighbor filter to achieve convolutional architecture is the graph convolutional network (GCN) \citep{kipf2017semi}. However, GCN merely considers the first-order neighbors, resting on which multiply layers are directly stacked to learn the multi-scale information, while it has been observed in many experiments that deeper GCN could not improve the performance and even performs worse \citep{kipf2017semi}. In other words, such convolutional filter limits the representational capacity of the model \citep{abu2019mixhop}.

In this work, we propose a new model, HWGCN, for convolutional filter formulation that is capable of mixing its neighborhood information at different orders to capture the expressive representations from the graph. Considering that convolution kernels of different sizes may extract different aspects or information from the input images, similarly, the size of convolutional filter plays a very important role for neighborhood mixing in graph convolutions. Researchers have recently made some attempts to deal with higher-order neighbors for the convolutional filter \citep{abu2019mixhop, liao2019lanczos}. 

Instead of using adjacency matrix power with potential information overlap at different orders, in our proposed graph model HWGCN, we bring an important insight to leverage node features in addition to graph structure for convolutional filter formulation, which allows a refined architecture to code better with neighbor selection at different distances, and thus learn better node representations from first-order and higher-order neighbors.

Our contributions are four-fold. Firstly, we analyze the GCN and demonstrate the importance of similarity between first-order and higher-order information. Secondly, we build the convolutional filters with first-order and higher-order neighbors rather than the local neighborhood considered in previous work. Thirdly, we leverage Lasso and the information of node features and graph structure to minimize the feature loss between the first-order and higher-order neighbors to effectively aggregate the higher-order information in a weighted, orthogonal, and unsupervised fashion, unlike existing models that merely utilize graph structure. Fourthly, we conduct comprehensive experimental studies on a number of datasets, which demonstrate that HWGCN can achieve the state-of-the-art results in terms of classification accuracy.

\section{Preliminaries}
Let $\mathcal{G}=(\mathcal{V}, \mathcal{E}, A)$ be a graph, where $\mathcal{V}$ is the set of vertices $\{v_1,\cdots,v_n\}$ with $|\mathcal{V}| = n$, $\mathcal{E}$ is the set of edges, and $A$ is its first-order neighbor matrix, also called the adjacency matrix, where $A \in \mathbb{R}^{n \times n} $ and $A_{ij} = \{0, 1\}$, \emph{i.e.}, if $(v_i, v_j) \in \mathcal{E}$, then $A_{ij} = 1$; otherwise, $A_{ij} = 0$. Based on the adjacency matrix $A$, the diagonal degree matrix $D$ can be defined as $D_{ii} = \sum_{j=1}^n A_{ij}$. The feature matrix $X$ is denoted as $X = [x_1, x_2,  \cdots, x_n]^T\in \mathbb{R}^{n \times c}$, where for each node $v \in \mathcal{V}$, its feature $x_v \in \mathbb{R}^{c_0}$ is a $c_0$-dimensional row vector. In addition, since the Graph Convolution Network (GCN) proposed by \citep{kipf2017semi} is exploited as a base model to facilitate the analysis and understanding of our further proposed approach, we would like to briefly present its architecture here. The graph convolutional layer is defined as:
\begin{equation}
H^{(i)} = f(H^{(i-1)}, A) = \sigma \left(\tilde{D}^{-\frac{1}{2}}\tilde{A}\tilde{D}^{-\frac{1}{2}}H^{(i-1)}\Theta^{(i)}\right),
\end{equation}
where $H^{(i-1)}$ and $H^{(i)}$ are the input and output activations for layer $i$ (i$\ge$1), $\tilde{A}$ is a symmetrically normalized adjacency matrix with self-connections $A+I$, $I$ is the identity matrix, $\tilde{D}$ is the diagonal degree matrix of $\tilde{A}$, $\sigma$ is the non-linear activation function (e.g., ReLU), and $\Theta^{(i)} \in \mathbb{R}^{c_{i-1} \times c_{i}}$ is the learnable weight matrix for layer $i$. Given $H^{(0)} = X$, the GCN model with $l$ layers can be thus defined as:
\begin{equation}
Z = \operatorname{softmax} \left(\tilde{D}^{-\frac{1}{2}}\tilde{A}\tilde{D}^{-\frac{1}{2}}\cdots\ \sigma\left(\tilde{D}^{-\frac{1}{2}}\tilde{A}\tilde{D}^{-\frac{1}{2}}X\Theta^{(1)}\right)\cdots\Theta^{(l)}\right).
\end{equation}
Here, we are interested in semi-supervised node classification tasks. To train a GCN model on such a task with $c_{l}$ class labels, the softmax function normalizes the final output matrix $Z \in \mathbb{R}^{n \times c_{l}}$, where each row represents the probability of $c_{l}$ labels for a node. The cross-entropy loss can be accordingly evaluated between all the $\ervz$s and the corresponding nodes with known labels and the weights can be calculated with back propagation using some gradient descent optimization algorithms (e.g., Adam \citep{kingma2015adam}).

\section{Proposed Method} \label{sec:filterformulation}
In this section, we discuss the details of how we present $k^{th}$-order adjacency matrices and weight matrices to leverage Lasso, node features and graph structure simultaneously for the convolutional filter formulation, and how the HWGCN model benefits from such elaborated filters with neighborhood information.

\subsection{higher-order adjacency matrix and weight matrix}

Formulating a convolutional filter should allow GCN models to learn the node representation differences among the neighbors at different orders. 
However, the GCN proposed by \citep{kipf2017semi} simplified the Graph Laplacian by restricting the filters to merely operate the first-order neighbors around each node, which fails to capture this kind of semantics, even when stacked over multiple layers \citep{abu2019mixhop, zhou2018graph}. To put it into perspective, we evaluate the performance of GCN of different layers on Cora, and the results are illustrated in Table~\ref{different layer gcn}. We can observe that deeper GCN models are unable to improve the performance of node classification and even harm the prediction. That is to say, stacking multiple layers with one-hop message propagation is not necessary to yield an advantage to learn the latent information from higher-order neighbors.

\begin{table}[htbp]
\vspace{-0.2cm}
	\centering
	\caption{Different layers of GCN}
	\begin{tabular}{cccc}
		\toprule  
		One-layer GCN & Two-layer GCN & Three-layer GCN & Four-layer GCN\\ 
		\midrule  
		71.9\% & 80.1\% & 77.0\% & 70.6\% \\
		\bottomrule  
	\end{tabular}
	\label{different layer gcn}
\end{table}

From the point of view of the convolution operations over the images, the filters of different neighborhood sizes generally contribute to greater flexibility and thus better performance on various computer vision tasks, while such approximations into graph domain have been scarce with some exceptions that few non-spectral approaches attempted to take advantage of larger yet fixed filter size \citep{atwood2016diffusion,niepert2016learning} which are somewhat unsatisfying for spectral filter extension; \citep{abu2019mixhop} designed MixHop to mix the feature representations of higher-order neighbors in one graph convolution layer. Since $A$ is a nonnegative matrix, if $A_{ij} > 0$, then its matrix power $A^k_{ij}$ would be positive, such that $A$ and $A^k$ have non-zero elements in the same positions of matrices. This implies, the layer output may impose the lower-order information on higher orders and increase the feature correlations. To further explain this, we present our Theorem~\ref{theorem:1} with proof and analysis left in Appendix~\ref{sec:appendix1}.

\begin{theorem}\label{theorem:1}
Let $A^{p}$ and $A^{q}$ denote the adjacency matrix $A$ multiplied by itself $p$ and $q$ times respectively where $p, q \in \mathbf{N}^{*}$ and $p < q$, then $A^{p} \circ A^{q} \ne 0$ if there are two walks between node $v_i$ and $v_j$ where the length of two walks are $p$ and $q$ respectively.
\end{theorem}

To this end, we would like to formulate a convolutional filter using first- and higher-order neighbors, so that the lower-order neighborhood information will be distinctively mixed with higher orders, while not overlapping from each other. We first introduce the concept of $k^{th}$-order adjacency matrix.

\begin{definition}
\textbf{$k^{th}$-order adjacency matrix}. Given a graph $\mathcal{G}=(\mathcal{V}, \mathcal{E}, A)$, we use the shortest path distance to determine the order between each pair of nodes. Let the shortest path distance between node $v_{i}$ and node $v_{j}$ be $d_{ij}$, and $k^{th}$-order adjacency matrix be $A^{(k)} \in \mathbb{R}^{n \times n} $, so that the element $A_{ij}^{(k)} = \{0, 1\}$, i.e., if $d_{ij} = k$, then $A_{ij}^{(k)} = 1$; otherwise, $A_{ij}^{(k)} = 0$.
\end{definition}
Based on the definition, the $k^{th}$-order adjacency matrix $A^{(k)}$ can be accordingly denoted as follows: 
\begin{equation}
\label{math:4}
A_{ij}^{(k)}=\begin{cases}
 1 &  \ \ d_{ij} = k \\
0 &   \ \ d_{ij} \not= k \end{cases}
\end{equation}

\begin{corollary}\label{corollary:1}
Let $A^{(p)}$ and $A^{(q)}$ are $p^{th}$-order and $q^{th}$-order adjacency matrices respectively where $p, q \in \mathbf{N}^{*}$ and $p \ne q$, then $A^{(p)} \circ A^{(q)} = 0$.
\end{corollary}

The proof of Corollary~\ref{corollary:1} can be found in Appendix~\ref{sec:appendix2}. Given the adjacency matrices of different orders, a naive solution to formulate the filter is to add all the $k^{th}$-order adjacency matrices $A^{(k)}$ $(1 < k \le K)$ to $A$.

\begin{figure}[htbp!]
	\centering
	\includegraphics[width=0.85\linewidth]{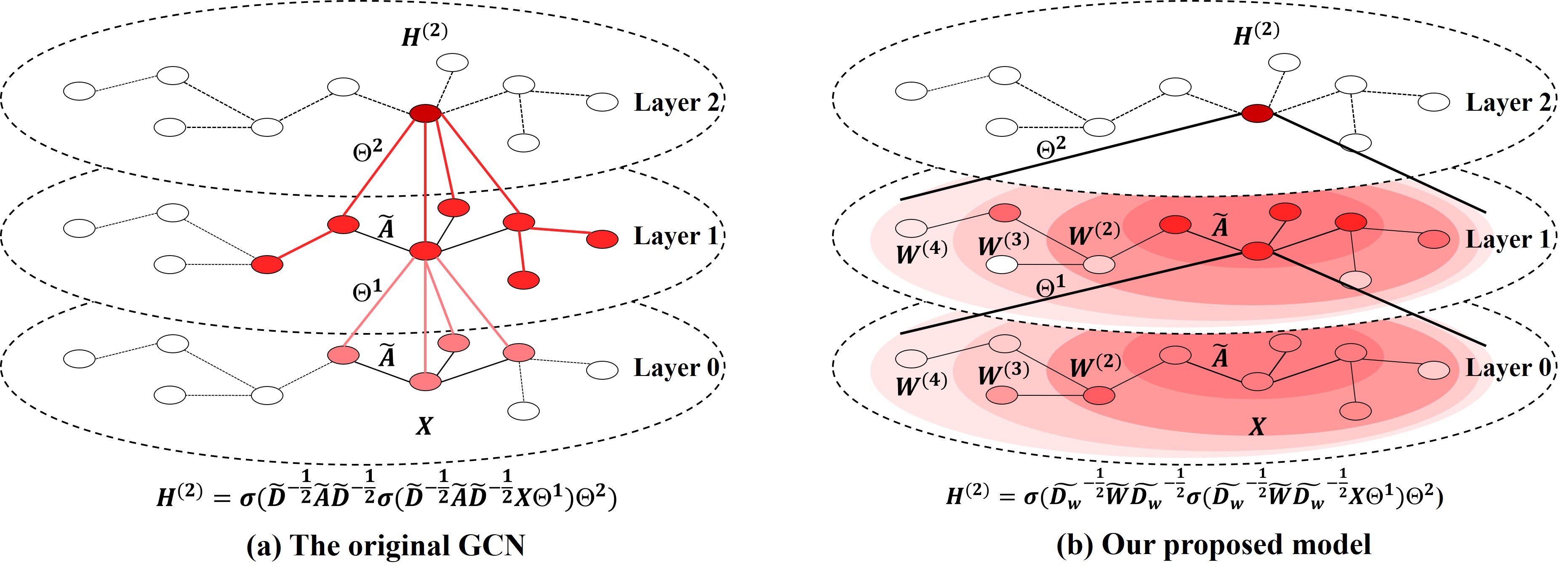}
	\caption{(a) First-order graph convolution presented in the original GCN (two layers) versus (b) our proposed convolution model propagating different-order neighborhood information (two layers).} \label{fig:comparison}
\end{figure}

However, this solution could generate an extremely dense matrix to propagate the noisy information from increasing number of expanded neighbors over layers, and yet make no distinction among neighbors at different orders. Note that, different  neighbors (\emph{i.e.}, at different orders or different positions in the same order) contribute to the node semantics differently. Thus, a more sophisticated solution to formulate the filter is 
to assign different weights to higher-order neighbors specifying their layer-wise and node-wise importances, so that each non-zero element in $k^{th}$-order adjacency matrix will be represented as a weight. Following $k^{th}$-order adjacency matrix's definition, we introduce $k^{th}$-order weight matrix as:
\begin{definition}
\textbf{$k^{th}$-order weight matrix}. Let the shortest path distance between node $v_{i}$ and node $v_{j}$ be $d_{ij}$, and $k^{th}$-order weight matrix be $W^{(k)} \in \mathbb{R}^{n \times n}$, such that the element $W_{ij}^{(k)} = \{0, w_{ij}^{(k)}\}$ $(w_{ij}^{(k)} \ge 0)$, i.e., if $d_{ij} = k$, then $W_{ij}^{(k)} = w_{ij}^{(k)}$; otherwise, $W_{ij}^{(k)} = 0$.

\end{definition}
Accordingly, the $k^{th}$-order weight matrix $W^{(k)}$ can be formulated as follows: 
\begin{equation}
W_{ij}^{(k)}=\begin{cases}
w^k_{ij}  &  \ \ d_{ij} = k \\
0 &   \ \ d_{ij} \not= k \end{cases}
\quad\quad s.t.\quad w^k_{ij} \geq 0.
\end{equation}
As such, $k^{th}$-order weight matrix has several significant properties: (1) $W^{(p)} \circ W^{(q)} = 0$ for any pair of $p^{th}$-order and $q^{th}$-order weight matrices; (2) $w^k_{ij}$ is learnable to specify the importance of each higher-order neighbor; (3) $W^{(k)}$ is more sparse than $A^{(k)}$ since $w^k_{ij}$ could be minimized to zero using some optimizations. Considering that $A$ approximates the most direct and effective layer-wise modeling capacity \citep{kipf2017semi}, we therefore propose to add all the $k^{th}$-order weight matrices $W^{(k)}$ $(2 < k \le K)$ to $A$, so that the graph Laplacian can be formed as $\tilde{D}_w^{-\frac{1}{2}}\tilde{W}\tilde{D}_w^{-\frac{1}{2}}$ where $W = A + \sum_{k=2}^KW^{(k)}$, $\tilde{W} = W + I$ and $\tilde{D}_w$ is the degree matrix of $\tilde{W}$. The difference between the first-order graph convolution and our proposed graph convolution model with two layers is shown in Figure~\ref{fig:comparison}. We will discuss how we adopt Lasso idea to determine the value of each element of the $k^{th}$-order weight matrix in section~\ref{subsec:weightmatrix}. 

\subsection{Learning Weight Matrices} \label{subsec:weightmatrix}

In statistics, feature selection is to select a subset of relevant explanatory variables to describe a response variable \citep{fonti2017feature}. Robert Tibshirani proposed the Lasso \citep{tibshirani1996regression} to perform this process using $\ell_{1}\text{-norm}$ which can shrink some coefficients and set others to zero. Inspired by its effectiveness on variable selection, here we apply Lasso for higher-order neighbor selection. 
As we can see, the first-order neighbors have a close relationship and share some significant commonalities with the central node, and thus have less noisy labels, which acts as a basic and important assumption to GCN. Therefore, we use the first-order neighbors as the observed features and extract the higher-order neighbors with feature vectors paralleled to the aggregated feature vectors of first-order neighbors in $n$-dimensional space. Accordingly we propose the following method:
\begin{equation}
 \mathcal{L}=\sum\limits_{i=1}^n\mathcal{L}_i=\sum\limits_{i=1}^n\left( \left\Vert\sum_{j=1}^n\left(x_j-\left(\tilde{S}X\right)_i\right)W_{ij}^{(k)}\right\Vert_2^2 + \lambda\sum_{j=1}^{n} |W_{ij}^{(k)}| \right) 
\end{equation}
where the affinity matrix $\tilde{S}
=\tilde{D}^{-\frac{1}{2}}\tilde{A}\tilde{D}^{-\frac{1}{2}}$. $\mathcal{L}_i$ can be transformed to an optimization problem as follows \citep{breiman1995better}:
\begin{equation}
\operatorname{minimize} \left\Vert\sum_{j=1}^n\left(x_j-\left(\tilde{S}X\right)_i\right)W_{ij}^{(k)}\right\Vert_2^2
\quad
\operatorname{subject   \ to} \sum_{j=1}^{n} |W_{ij}^{(k)}| \leq t
\end{equation}
$\lambda$ is the parameter that controls the strength of the penalty, where the larger the penalty value of $\lambda$, the greater the amount of shrinkage. When $\lambda$ goes to infinity, $t$ becomes 0 and all coefficients shrink to 0. Note that, the sum of coefficients $\sum_{j=1}^{n} |W_{ij}^{(k)}|$ controls the potential scale of the feature vector, and thus we need to effectively restrict its value. Due to its penalty nature, we can set $\lambda$ to a large value and shift the sum (\emph{i.e.}, $\sum_{j=1}^{n} |W_{ij}^{(k)}|$) to a constant value $\alpha$ instead of the cross validation for $\lambda$. In this respect, we can obtain the $l_1$ optimization problem as follows:

\begin{equation}
\operatorname{minimize}  \left\Vert\sum_{j=1}^n\left(x_j-\left(\tilde{S}X\right)_i\right)W_{ij}^{(k)}\right\Vert_2^2
\quad
\operatorname{subject   \ to} \ W_{ij}^{(k)} \geq 0, \ \sum_{j=1}^{n} W_{ij}^{(k)} = \alpha_{i} \ 
\end{equation}
where $\alpha_{i}$ is the sum of the scale coefficients for $k^{th}$-order neighbors of node $i$. 

Considering that the number of higher-order neighbors may be different from that of first-order neighbors, it is necessary for us to perform scale transformation for the higher-order neighbors. Given the scale of the aggregated feature vector of first-order neighbors, we can naturally leverage it to control the scale of feature vector for higher-order neighbors. To this end, we propose the following loss function:
\begin{equation}
	\mathcal{L}_{norm}= \left\Vert D_{\alpha}^{(k)}A^{(k)}X-\tilde{S}X\right\Vert_{F}^{2} = \sum_{i=1}^n \mathcal{L}^{(i)}_{norm} =\sum_{i=1}^n \left\Vert\alpha_i^{(k)}\sum\limits_{j=1}^nA_{ij}^{(k)}x_j-\left(\tilde{S}X\right)_i\right\Vert_2^2 \\
\end{equation}
where $D_{\alpha}^{(k)} = \mathrm{diag}\left\{{\alpha_1^{(k)}} {\dots} \alpha_n^{(k)}\right\}$ and $\alpha_i^{(k)}$ is the scale coefficient for node $i$. By solving $\frac{d \mathcal{L}_{norm}^{(i)}}{d \alpha_i^{(i)}}=0$, we can get the approximate solution of $\alpha_i^{(k)}$ as follows:

\begin{equation}
\alpha_i^{(k)} = \frac{\left(\sum_{j=1}^n A_{ij}^{(k)}x_j\right)\left(\tilde{S}X\right)_i^T}{\left(\sum_{j=1}^n A_{ij}^{(k)}x_j\right)\left(\sum_{j=1}^n A_{ij}^{(k)}x_j\right)^T}
\end{equation}

Accordingly, $\mathcal{L}_i$ has the following form:
\begin{equation}
    \begin{cases}
     \mathcal{L}_i=\left\Vert \sum\limits_{j=1}^n\left(x_j-\left(\tilde{S}X\right)_i\right)W_{ij}^{(k)}\right\Vert_2^2
 \\\sum\limits_{j=1}^nW_{ij}^{(k)} = \sum\limits_{j=1}^n\left(D_{\alpha}^{(k)} A_k\right)_{ij}\\W_{ij}^{(k)} \geq 0
  \end{cases}
\end{equation}

\citep{osqp} proposed the OSQP which is a fast method \citep{osqp-codegen,osqp-infeasibility} based on ADMM (Alternating Direction Method of Multipliers) to solve this quadratic problem. We can assign different weights to different neighbors by solving this optimization problem. Since the Pubmed dataset in table \ref{statistics of the datasets} is a large graph and each node has too many higher-order neighbors that may mix features from unrelated neighbors, we sort by the weight of the neighbor nodes, and select a certain proportion of neighbor nodes to form the final neighbor weight matrix. This may improve the performance and reduce training time. The proportion of neighbor nodes for Pubmed is as follows:

\begin{table}[htbp]
\normalsize
    \setlength{\tabcolsep}{5mm}
	\centering
	\caption{Reserved proportion for Pubmed}
	\label{pubmed proportion}
	\begin{tabular}{ccccc}
	\toprule
		\textbf{Order} & $2^{nd}$ & $3^{rd}$ & $4^{th}$ & $5^{th}$\\
		\toprule  
		 \textbf{Proportion} & 20\% & 10\%  & 5\% & 5\% \\ 
    \bottomrule
	\end{tabular}
\end{table}

The statistics of $W^{(k)}$ is displayed in Appendix \ref{sec:statistics}. After obtaining $k^{th}$-order weight matrix $W^{(k)}$, we replace the adjacency matrix $A$ in the original GCN with $W$ and formulate our model HWGCN as:
\begin{equation}
Z = \operatorname{softmax} \left(\tilde{D}_w^{-\frac{1}{2}}\tilde{W}\tilde{D}_w^{-\frac{1}{2}}\cdots\ \sigma\left(\tilde{D}_w^{-\frac{1}{2}}\tilde{W}\tilde{D}_w^{-\frac{1}{2}}X\Theta^{(1)}\right)\cdots\Theta^{(l)}\right).
\end{equation}

\section{Experiments}
In this section, we evaluate the performance of our proposed model HWGCN on a number of datasets. We perform experiments on the graph-based benchmark node classification tasks with the random splits and fixed split \citep{yang2016revisiting} of each dataset, and compare our model with a wide variety of previous approaches and state-of-the-art baselines. 

\subsection{Datasets}
We test our model on three citation network benchmark datasets: \emph{Cora}, \emph{Citeseer} and \emph{Pubmed} \citep{yang2016revisiting,kipf2017semi} - in all of these datasets, nodes represent documents and edges denote citation links; node features correspond to elements of a bag-of-words representation of a document \emph{i.e.}, 0/1 values indicating the absence/presence of a certain word, while each node has a class label \citep{velickovic2018graph}. The dataset statistics are summarized in Table \ref{statistics of the datasets}. 

\begin{table}[htbp]
\vspace{-0.3cm}
\normalsize
    \setlength{\tabcolsep}{5mm}
	\centering
	\caption{Statistics of the datasets used in our experiments}
	\label{statistics of the datasets}
	\begin{tabular}{ccccc}
	\toprule
		\textbf{Dataset} & \textbf{Nodes} & \textbf{Edges} & \textbf{Classes} & \textbf{Features}\\
		\toprule  
		Cora & 2,708 & 5,429  & 7 & 1,433 \\ 
        Citeseer & 3,327 & 4,732  & 6 & 3,703 \\
        Pubmed & 19,717 & 44,338  & 3 & 500 \\
    \bottomrule
	\end{tabular}
\vspace{-0.3cm}
\end{table}

\subsection{Baselines}
We compare our approach against some previous methods and state-of-the-art baselines, including: label propagation (LP) \citep{zhu2003semi}, semi-supervised embedding (SemiEmb) \citep{weston2012deep}, manifold regularization (ManiReg) \citep{belkin2006manifold}, skip-gram based graph embeddings (DeepWalk) \citep{perozzi2014deepwalk}, iterative classification algorithm (ICA) \citep{lu2003link}, and Planteoid \citep{yang2016revisiting} on fixed-split datasets; multi-layer perceptron, \emph{i.e.}, MLP (without adjacency matrix), graph attention networks (GAT) \citep{velickovic2018graph}, plain GCN \citep{kipf2017semi} and MixHop \citep{abu2019mixhop} on both fixed-split and random-split datasets.

\subsection{Experimental Setup} \label{setup}

We closely follow the experimental setup in \citep{kipf2017semi} to implement our model for evaluation. The model HWGCN, a two-layer GCN structure with 16 hidden units, is trained through 200 maximum epochs using Adam \citep{kingma2015adam} with 0.01 initial learning rate, 5 $\times$ $10^{-4}$ L2 regularization on the weights, and 0.5 dropout rate for the input and hidden layers. The parameter settings of GAT and MixHop are directly taken from \citep{velickovic2018graph} and \citep{abu2019mixhop}. For datasets, we adjust the random splits of Cora, Citeseer, and Pubmed respectively to align with three different scenarios: 5, 10 or 20 instances for each class are randomly sampled as training data while another 500 and 1000 instances are selected as validation and test data.  

\begin{figure}[htbp]
\centering
\subfigure{
\begin{minipage}[t]{0.30\linewidth}
\centering
\includegraphics[width=\linewidth]{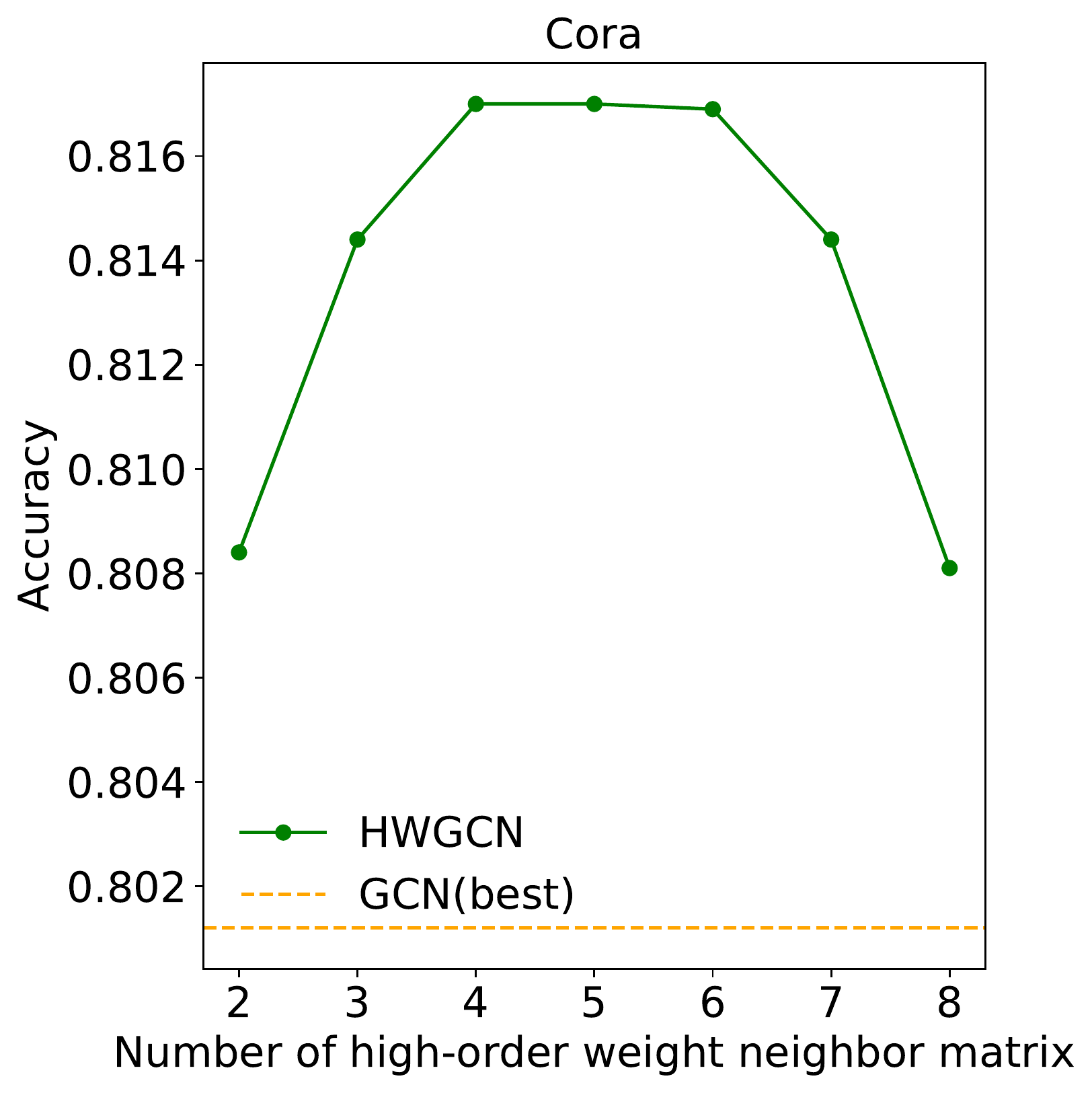}
\end{minipage}
}
\subfigure{
\begin{minipage}[t]{0.30\linewidth}
\centering
\includegraphics[width=\linewidth]{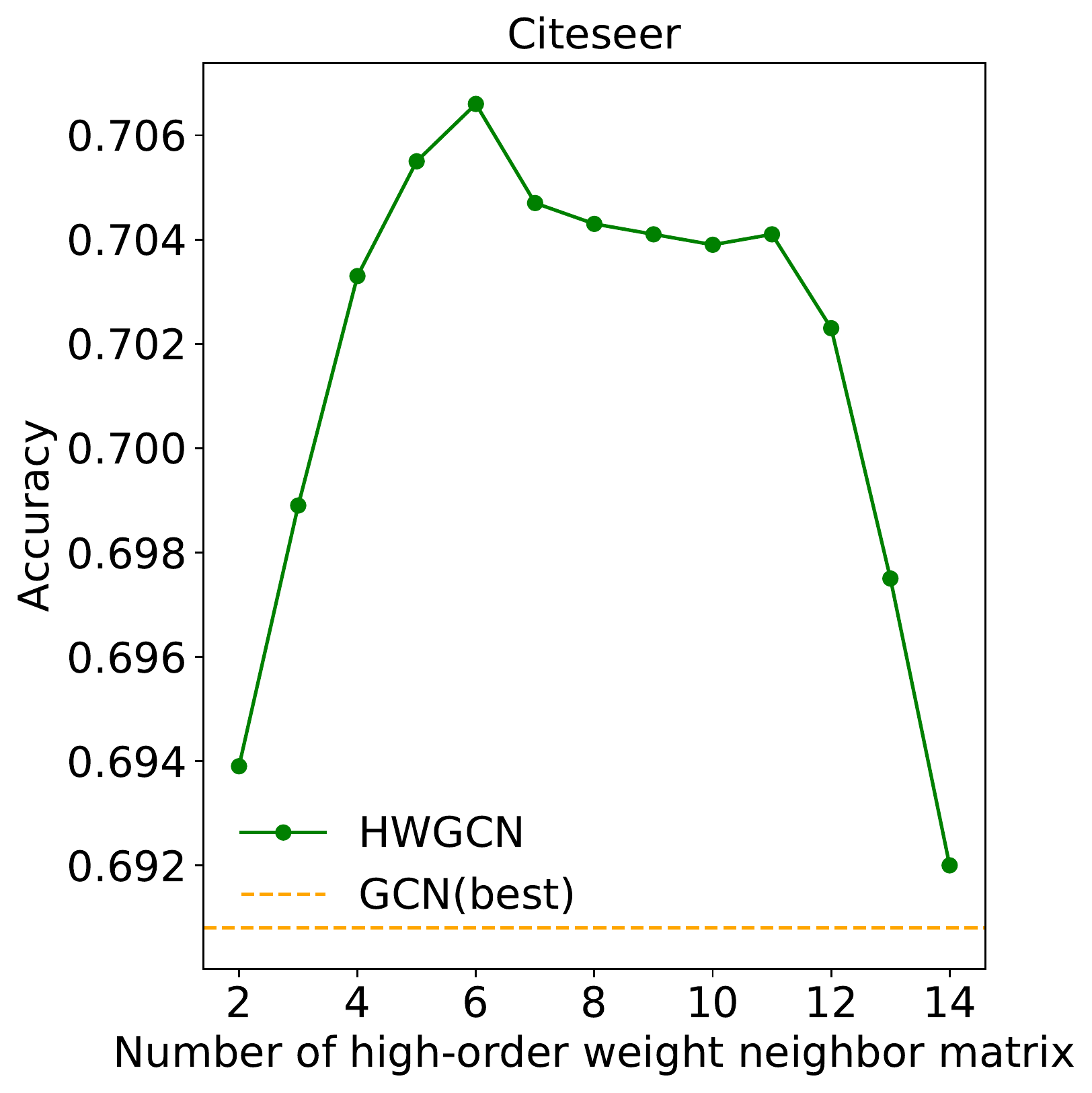}
\end{minipage}
}
\subfigure{
\begin{minipage}[t]{0.30\linewidth}
\centering
\includegraphics[width=\linewidth]{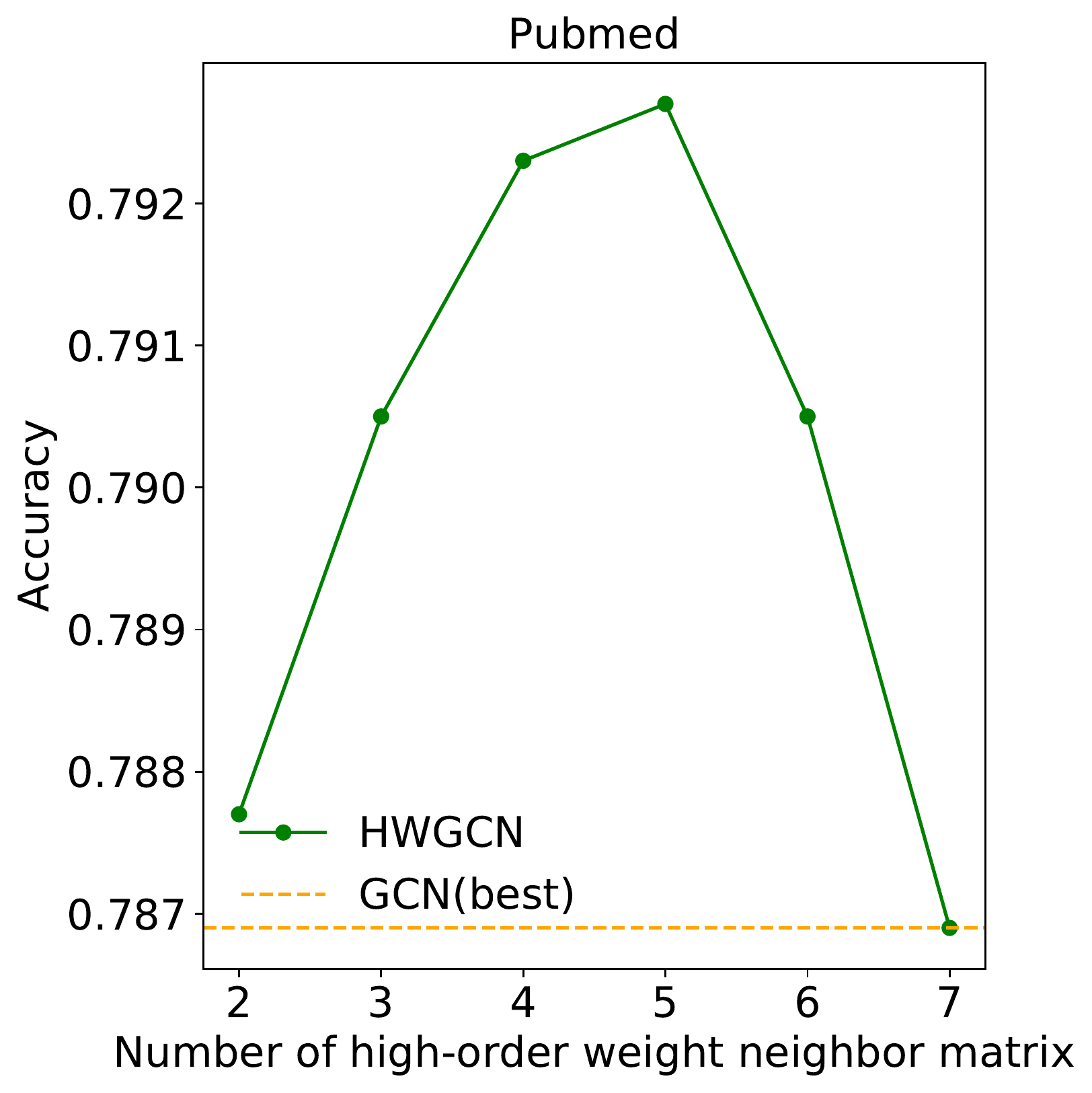}
\end{minipage}
}

\caption{Influence of neighbor matrix on classification performance.}
\label{train 20}
\vspace{-0.3cm}
\end{figure}

In addition, in order to fairly assess the benefits of our convolutional filter formulation mechanism with higher-order neighbor information, and to achieve the best performance, we further compare the results by increasingly adding different numbers of weight matrices $W^{(k)}$ to $A$ (\emph{i.e.}, $A + \sum_{j = 2}^{k}W^{(j)}$ \emph{s.t.} $k \ge 2$) to determine the effective neighborhood distance. Accordingly, we conduct such experiments using 20 instances for each class as training data and 1000 instances as test data. The results are illustrated in Figure \ref{train 20}, from which we can observe that, mixing neighborhood information at different distances in a graph show different performances for node classification; we obtain the best results with adding weight matrices to $k \in \{4, 5, 6\}$ for Cora, $k=6$ for Citeseer and $k = 5$ for Pubmed respectively. Based on this observation, we therefore set $k = 6$ for Cora, $k = 6$ for Citeseer, and $k = 5$ for Pubmed in the following experiments.

\subsection{Results}
The results of our comparative evaluation experiments on random-split datasets are summarized in Table \ref{random splits}. We report the mean classification accuracy on the test nodes of our method after 50 runs over random dataset splits with 5, 10 or 20 labels for per class. 

\begin{table}[htbp]
\normalsize
\centering
\caption{Summary of results on random-split datasets in terms of classification accuracy (\%)}
\setlength{\tabcolsep}{1.5mm}
\label{random splits}
\begin{tabular}{cccccccccc}
\toprule
\multirow{2}{*}{Training size} & \multicolumn{3}{c}{$20$} & \multicolumn{3}{c}{$10$} & \multicolumn{3}{c}{$5$} \\
\cmidrule(r){2-4} \cmidrule(r){5-7} \cmidrule(r){8-10}
&  Cora     &  Citeseer  &   Pubmed
&  Cora     &  Citeseer  &   Pubmed
&  Cora     &  Citeseer  &   Pubmed  \\
\midrule
MLP         &55.0  &55.5  &70.5   &48.0  &47.4  &64.9  &39.4  &40.0  &60.6 \\
GAT         &81.6  &70.1  &77.7   &78.9  &66.5  &74.1  &73.5  &61.3  &69.3  \\
GCN         &80.1  &69.1  &78.7   &76.6  &65.1  &74.6  &69.2  &59.1  &70.5  \\
MixHop      &80.5  &69.8  &\textbf{79.3}  &78.2  &66.8  &\textbf{75.9}  &74.0  &\textbf{62.6}  &70.5  \\
\midrule
HWGCN    &\textbf{81.7}  &\textbf{70.7}  &\textbf{79.3}  &\textbf{79.8}  &\textbf{67.2}  &75.8 &\textbf{74.6}  &61.9 &\textbf{71.7} \\
\bottomrule
\end{tabular}
\end{table}

From the results, we can see that our proposed model HWGCN successfully outperforms previous approaches and state-of-the-art baselines in most cases on random splits, where the best result for each column has been highlighted in bold. More specifically, compared to the best performances from MLP, GAT, GCN and MixHop, HWGCN manages to improve the accuracy by a margin of 0.1\% to 0.9\% on Cora with respect to different data splits. For Citeseer, and Pubmed which is a larger graph with more nodes, though MixHop that mixes neighborhood information at different distance performs slightly better when the training size is 5 per class (on Citeseer) and 10 per class (on Pubmed), HWGCN is still able to outperform the spectral methods such as GCN and GAT that limit to operating
first-order neighborhood. It is worth noting that we do not complicate our model to achieve such superior performance, which is trained as the same architecture as GCN. The success of our model lies in the proper consideration and accommodation of higher-order neighborhood information, and the advantage of weight matrix formulation to decrease the noises and thus improve the expressiveness of node representations. 

\begin{table}[htbp]
\normalsize
\centering
\caption{Classification results on fixed split (\%)}
\setlength{\tabcolsep}{7mm}
\label{fixed split}
\begin{tabular}{cccccccccc}
\toprule
\textbf{Method}&  \textbf{Cora}     &  \textbf{Citeseer}  &   \textbf{Pubmed}  \\
\midrule
MLP         &55.1  &46.5  &71.4   \\
ManiReg     &59.5  &60.1  &70.7   \\
SemiEmb     &59.0  &59.6  &71.7  \\
LP          &45.3  &68.0  &63.0  \\
DeepWalk    &67.2  &43.2  &65.3   \\
ICA         &75.1  &69.1  &73.9   \\
Planteoid   &75.7  &64.7  &77.2   \\
GCN         &81.5  &70.3  &79.0   \\
GAT         &\textbf{83.0}  &\textbf{72.5}  &79.0   \\
MixHop      &81.9  &71.4  &\textbf{80.8}   \\
\midrule
HWGCN    &82.9  &71.7  &80.5\\
\bottomrule
\end{tabular}
\end{table}

We also conduct experiments on fixed split and report the mean accuracy of 100 runs. Note that, for comparison purposes, we directly take the results of previous approaches including ManiReg, SemiEmb, LP, DeepWalk, ICA, Planteoid and MLP already reported in the original GCN paper \citep{kipf2017semi} and GAT from paper \citep{velickovic2018graph}. The results are summarized in Table \ref{fixed split}. Though it slightly falls behind GAT on Cora and Citeseer, and Mixhop on Pubmed, HWGCN still achieves state-of-the-art performance, which is better than GCN. Furthermore, based on the comprehensive results in Table~\ref{random splits} and \ref{fixed split}, we can see that our model using Lasso to select relevant higher-order neighbors is beneficial to alleviate the problem of overfitting.

\section{Related Work}

For non-spectral approaches that generalize convolutions to the graph domain, convolutions are directly defined on the graph, operating on spatially close neighbors. 
\citep{duvenaud2015convolutional} proposed convolutional networks to learn molecular fingerprints where information flows between neighbors in the graph. \citep{atwood2016diffusion} proposed the diffusion-convolution neural networks (DCNNs), which propagates features based on the transition matrix power series. Both approaches use a different number of neighbors among all nodes to extract local features. By contrast, \citep{niepert2016learning} extracted local features using fixed number of neighbors for each of the nodes, while \citep{monti2017geometric} proposed a unified framework allowing to generalize CNN architectures to non-Euclidean domains. In addition, some researchers applied the pooling operation on graph. For example, \citep{zhang2018end} designed a novel SortPooling layer to filter the outputs of the convolutional layer, and \citep{gao2019graph} also proposed novel graph pooling and unpooling, which allows neighbor selection for the central node. The aforementioned approaches define the convolution and pooling by performing aggregation and filtering respectively over the neighbors of each node, yielding impressive performance on node, link and graph classification.

Some researchers also define the convolution operations based on the spectral formulation, which are called spectral approaches. \citep{bruna2014spectral} generalized the convolution operator in the non-Euclidean domain by exploiting the spectrum of the graph Laplacian; this approach involves intense computation and uses non-spatially localized filters. \citep{defferrard2016convolutional} proposed to approximate the filters by computing the Chebyshev polynomial recurrently, generating fast and localized spectral filters.  \citep{kipf2017semi} further introduced the graph convolutional network (GCN) via limiting the spectral filters to first-order neighbors for each node. As mentioned earlier, GCN is the model on which our work is based. The most relevant work to our approach is MixHop \citep{abu2019mixhop}, which repeatedly mixed feature representations of neighbors at various distances. The major distinction is that we formulate the convolutional filter in a more sophisticated way to leverage node features and graph structure from higher-order neighbors while avoiding potential neighborhood information overlaps, as has been analyzed in more details in Section \ref{sec:filterformulation}.

\section{Conclusion}

The original GCN updates the state of nodes by the aggregation of feature information from directly neighboring nodes in every convolutional layer, but fails to learn the higher-order neighborhood information through operating multiply layers; its performance suffers a drop-off when it adjusts the number of layers over two. To address this, some recent research efforts have been conducted on mixing neighborhood information at different distances to improve the expressive power of graph convolutions, which are promising yet limiting to adjacency matrix power. In this paper, we propose a novel model HWGCN to formulate the convolutional filter to regularize first-order and higher-order neighbors in a weighted and orthogonal fashion, where node features and graph structure are leveraged to minimize feature loss through Lasso, extract relevant higher-order neighborhood information, and thus learn better node representations. Our method is a generic framework which can be further applied to various graph convolution network models. The experimental results based on the three standard citation network benchmark dataset demonstrate state-of-the-art performance being achieved, which match or outperform other baselines.

\bibliography{paper}

\newpage
\appendix
\section{Proof and Analysis relevant to matrix powers}\label{sec:appendix1}
To prove Theorem~\ref{theorem:1}, We first give Lemma~\ref{lemma:1} with proof as follows:
\begin{lemma}
\label{lemma:1}
Let $A$ be the adjacency matrix of a graph $\mathcal{G}=(\mathcal{V}, \mathcal{E})$. Then $\left(A^k\right)_{ij}$ is the number of walks from $v_i$ to $v_j$ in $\mathcal{G}$ of length k where $1 \leq i \leq j \leq |\mathcal{V}|$.
\end{lemma}

\begin{proof}
We prove it by induction. When $k=1$, if $A_{ij}=1$, there is a walk from $v_i$ to $v_j$ of length 1. We assume by induction that $(A^{k-1})_{ij}$ is the number of walks from $v_i$ to $v_j$ of length $k-1$, and we have $A^k$ = $A^{k-1}A$,  \emph{i.e.}, $(A^k)_{ij} = \sum_{m=1}^{|\mathcal{V}|}(A^{k-1})_{im}A_{mj}$.

Based on the induction hypothesis, $(A^{k-1})_{im}$ is the number of walks of length $k-1$ between $v_i$ and $v_m$. $A_{mj}=1$ if $m$ and $j$ connect, and $(A^{k-1})_{im}A_{mj}$ is the number of walks from $v_i$ to $v_j$ of length $k$ with $v_m$ as their penultimate vertex. 
By summing the number of walks with all vertices in graph $\mathcal{G}$ as their penultimate vertex, we can accordingly get the number of walks from $v_i$ to $v_j$.
\end{proof} 

As we have proved Lemma \ref{lemma:1}, we further provide the proof of Theorem \ref{theorem:1} as follows:
\begin{proof}
Since $\left(A^k\right)_{ij}$ is the number of walks from $v_i$ to $v_j$ in $\mathcal{G}$ of length $k$ as stated in Lemma~\ref{lemma:1}, if there are two walks between node $v_i$ and $v_j$ where the length of two walks are $p$ and $q$ respectively, then $(A^p)_{ij} > 0$ and $(A^q)_{ij} > 0$. Therefore, we have $(A^p)_{ij}(A^q)_{ij}>0$ and $A^p \circ A^q \ne 0$.
\end{proof}

\begin{figure}[htbp!]
	\centering
	\includegraphics[width=0.5\linewidth]{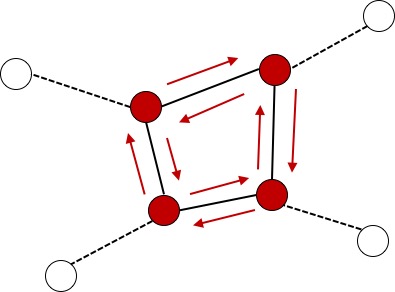}
	\caption{A cyclic graph example.} \label{fig:circle}
\end{figure}

By the theory, we can find that different matrix powers may have non-zero elements in the same position of matrix. Especially when there is a circle in graph as shown in Figure~\ref{fig:circle}, the corresponding elements appear cyclically in higher-order matrices. This accordingly results in information overlap between lower-order matrix and higher-order matrix. 

\section{Proof of Corollary ~\ref{corollary:1}}\label{sec:appendix2}
\begin{proof}
We can prove this corollary by two cases: first, for any node $v_i$ and $v_j$, if the shortest path distance between them are not $p$ or $q$, then $A_{ij}^{(p)} = 0$ and $A_{ij}^{(q)} = 0$; second, if the shortest path distance between them is $p$, then $A_{ij}^{(p)} = 1$. Since The shortest path distance between $v_i$ and $v_j$ has been uniquely determined, obviously $A_{ij}^{(q)} = 0$. For both cases, we have $A_{ij}^{(p)}A_{ij}^{(q)} = 0$ for any nodes $v_i$ and $v_j$, and thus $A^{(p)} \circ A^{(q)} = 0$.
\end{proof}

\newpage

\section{Accuracy curve for another two random splits and fixed split}
\begin{figure}[htbp]
\centering
\subfigure{
\begin{minipage}[t]{0.31\linewidth}
\centering
\includegraphics[width=\linewidth]{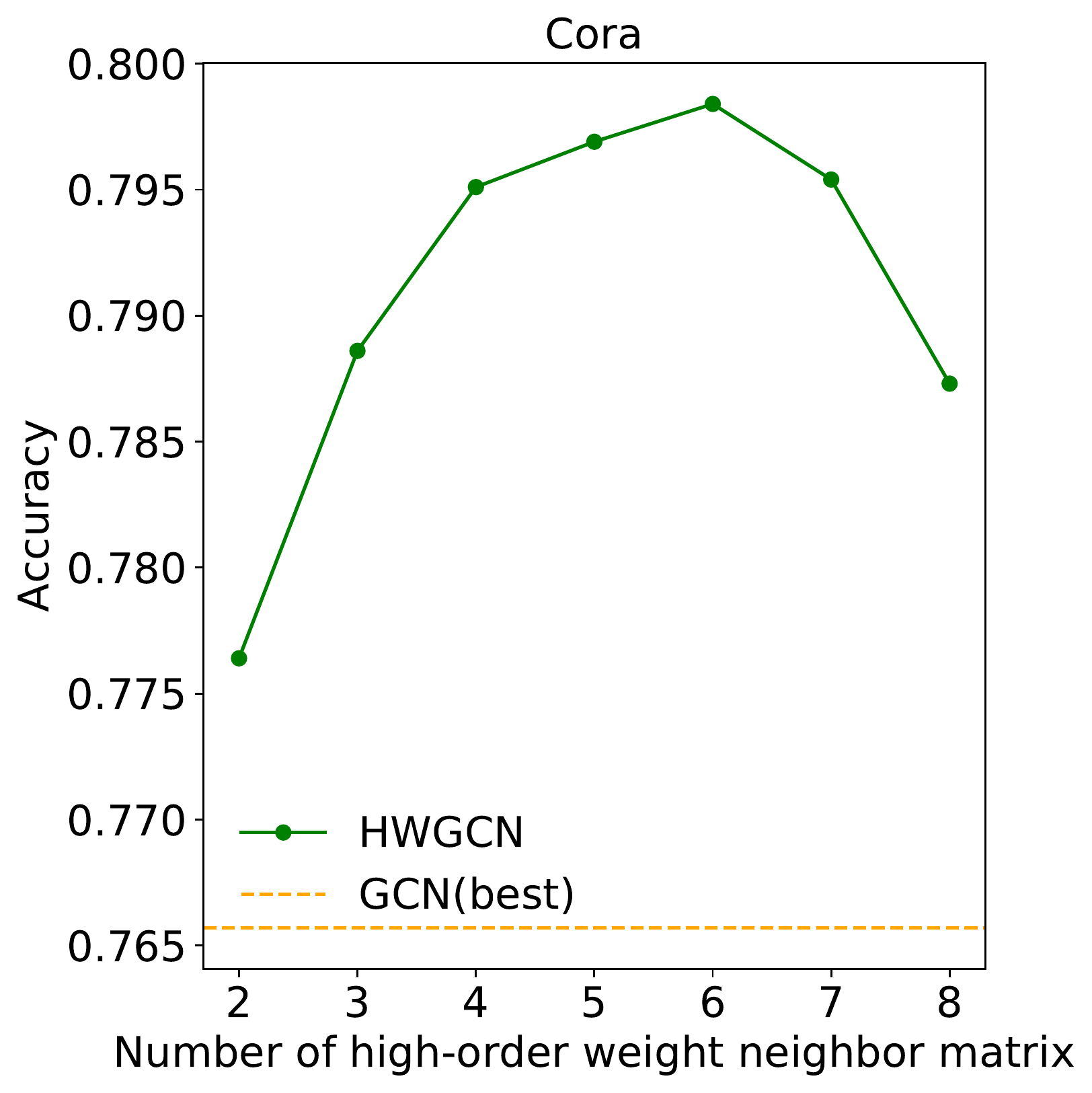}
\end{minipage}
}
\subfigure{
\begin{minipage}[t]{0.31\linewidth}
\centering
\includegraphics[width=\linewidth]{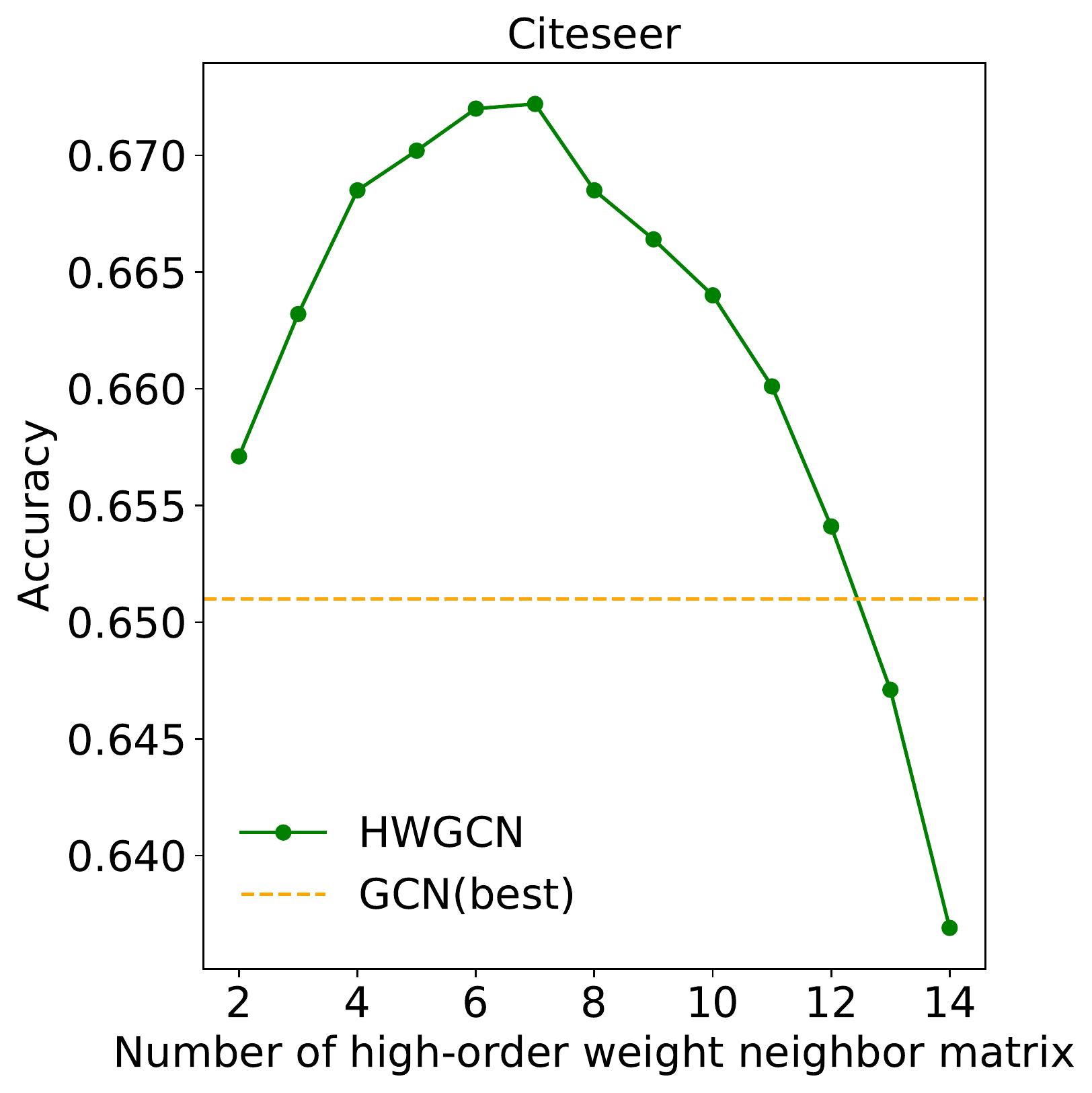}
\end{minipage}
}
\subfigure{
\begin{minipage}[t]{0.31\linewidth}
\centering
\includegraphics[width=\linewidth]{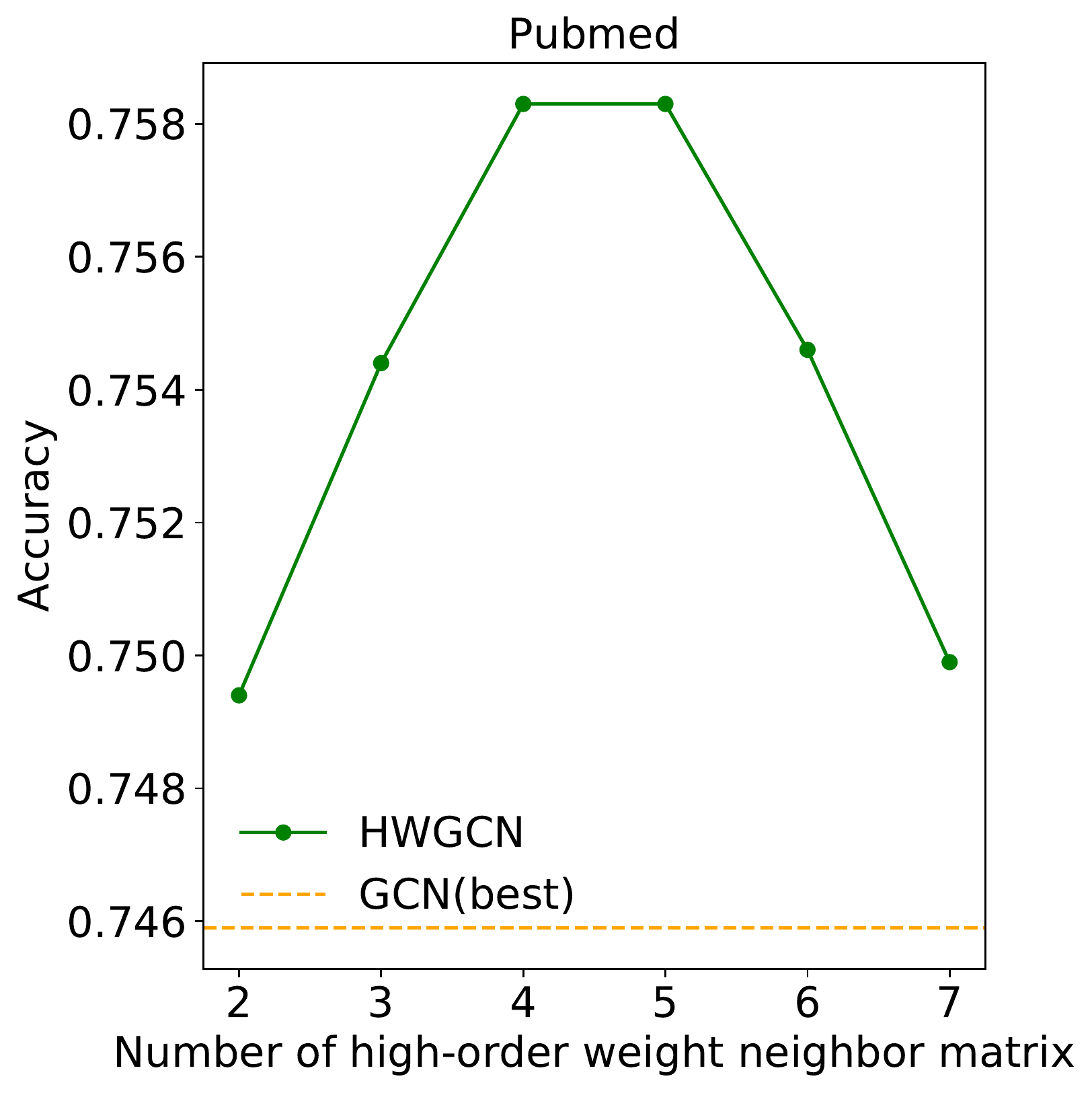}
\end{minipage}
}

\caption{Influence of neighbor matrix on classification performance (random split with 10 per class).}
\label{train 10}
\end{figure}

\begin{figure}[htbp]
\centering
\subfigure{
\begin{minipage}[t]{0.31\linewidth}
\centering
\includegraphics[width=\linewidth]{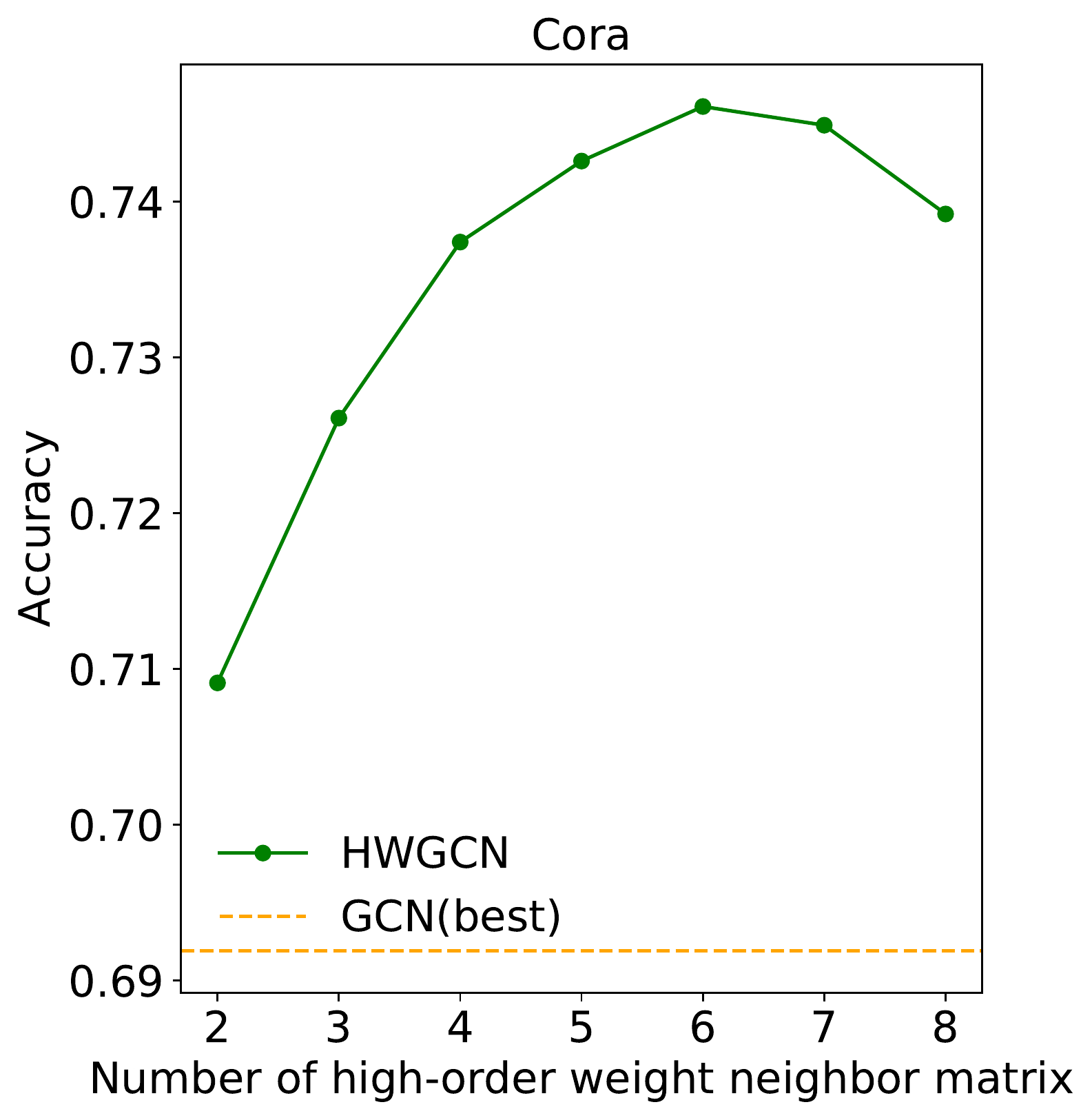}
\end{minipage}
}
\subfigure{
\begin{minipage}[t]{0.31\linewidth}
\centering
\includegraphics[width=\linewidth]{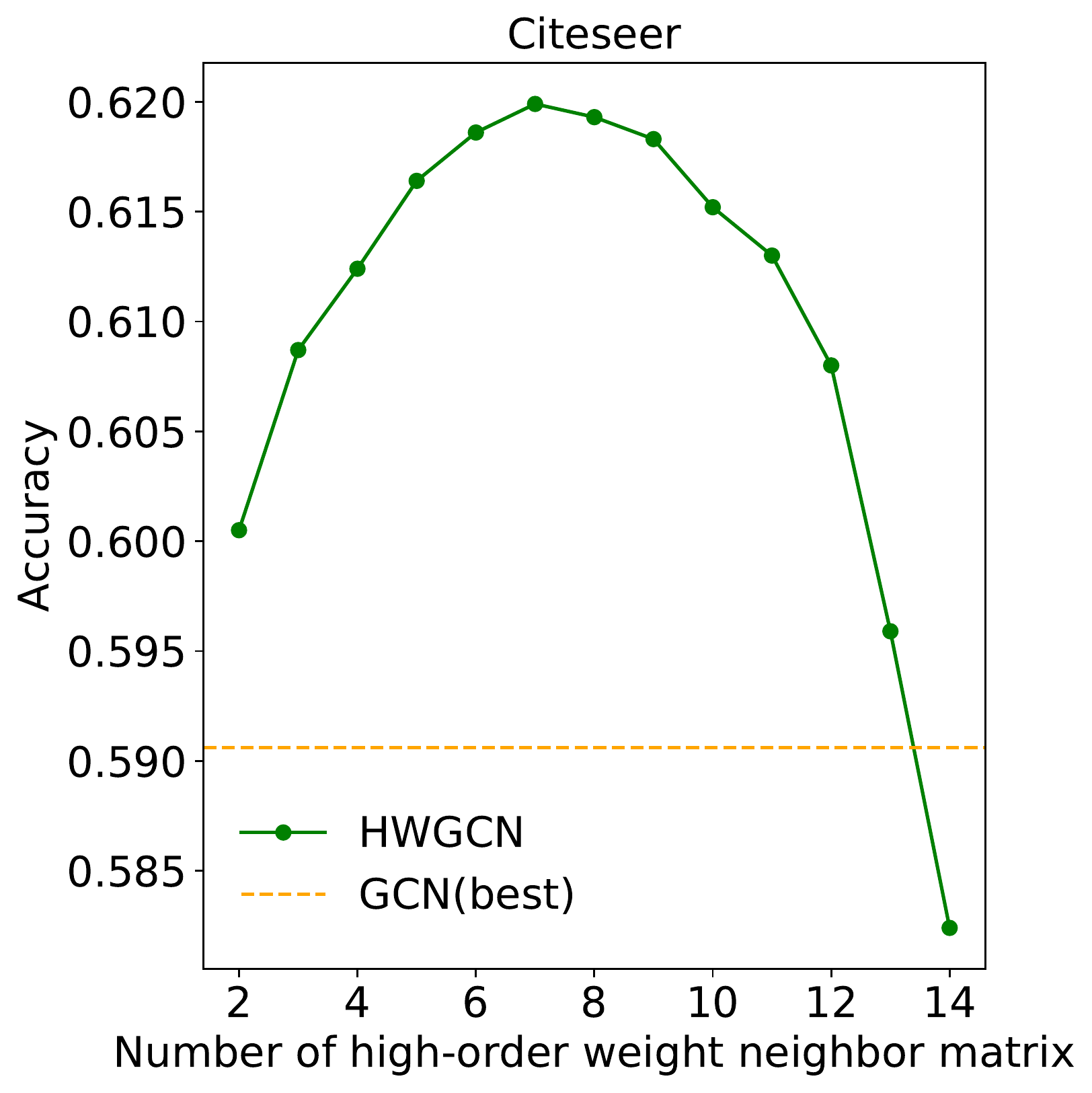}
\end{minipage}
}
\subfigure{
\begin{minipage}[t]{0.31\linewidth}
\centering
\includegraphics[width=\linewidth]{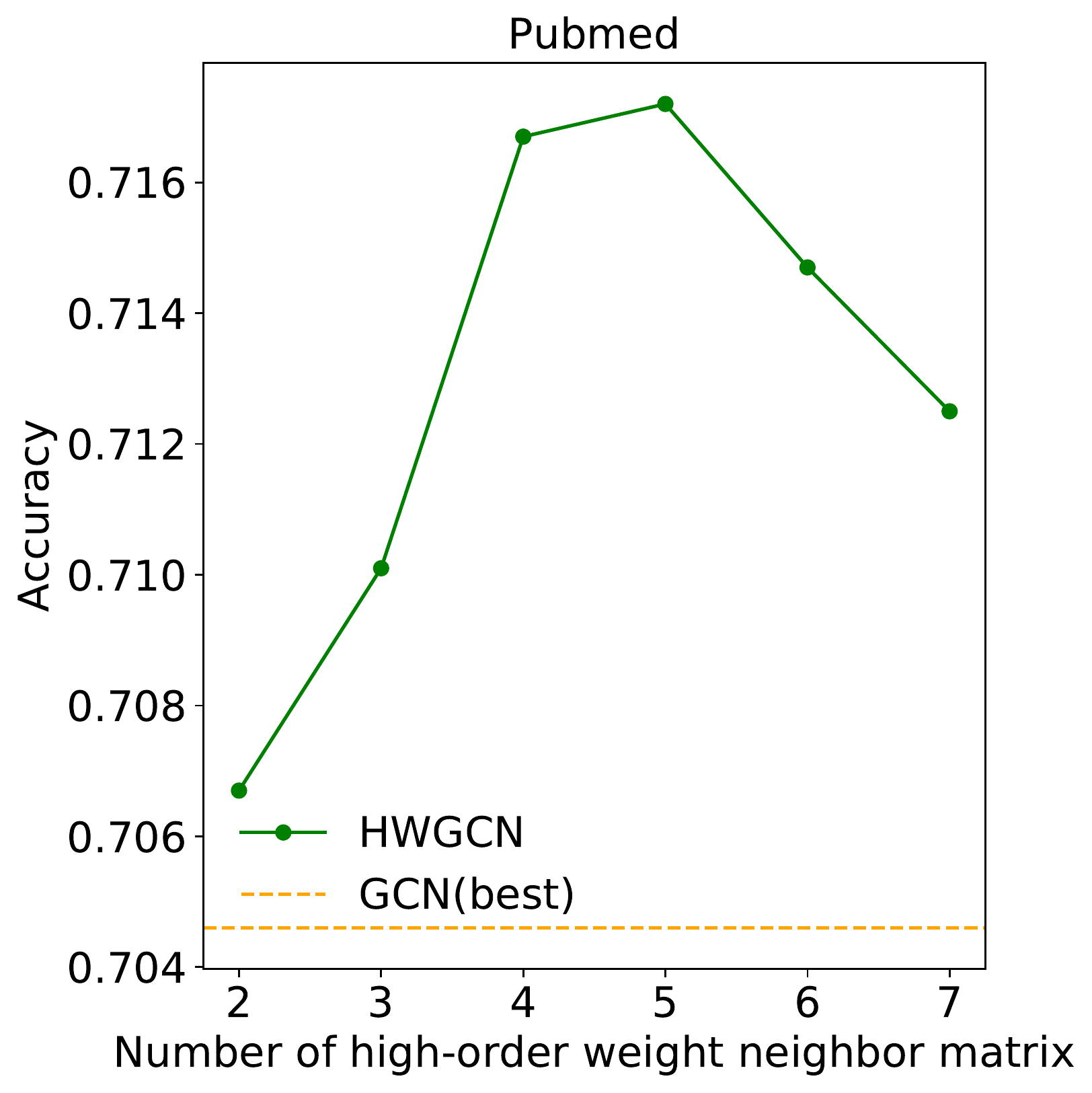}
\end{minipage}
}

\caption{Influence of neighbor matrix on classification performance (random split with 5 per class).}
\label{train 5}
\end{figure}

\begin{figure}[htbp]
\centering
\subfigure{
\begin{minipage}[t]{0.31\linewidth}
\centering
\includegraphics[width=\linewidth]{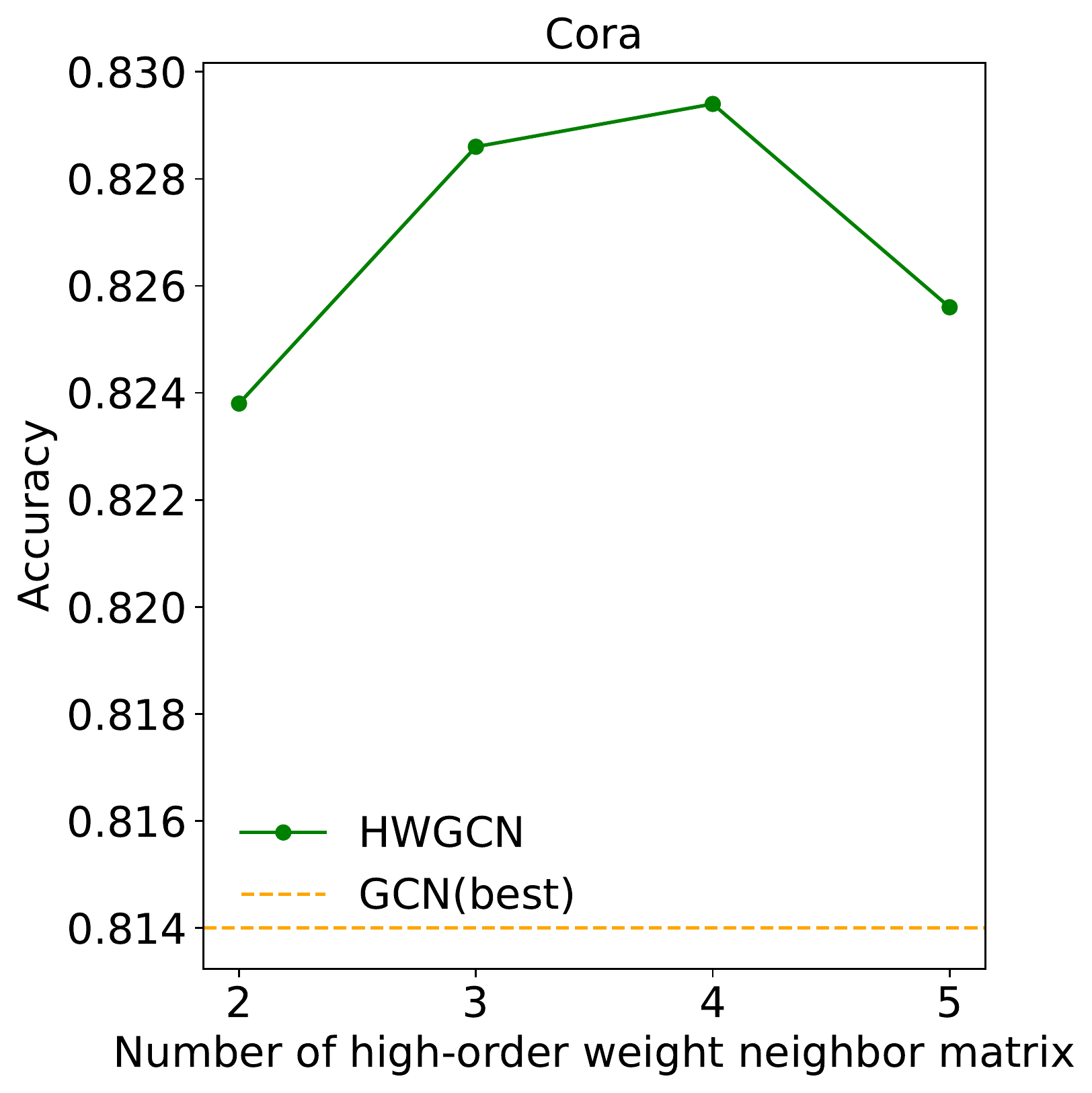}
\end{minipage}
}
\subfigure{
\begin{minipage}[t]{0.31\linewidth}
\centering
\includegraphics[width=\linewidth]{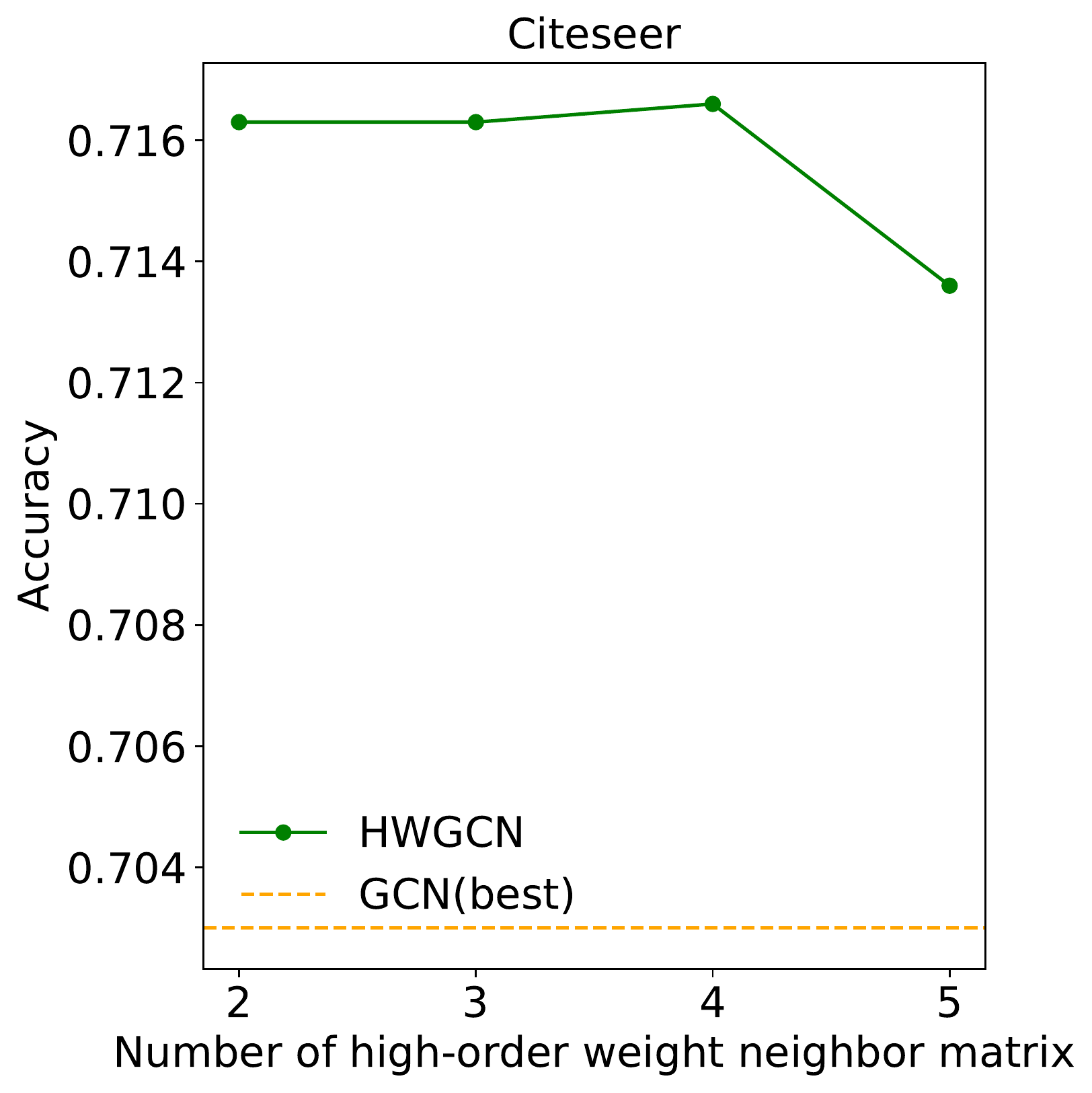}
\end{minipage}
}
\subfigure{
\begin{minipage}[t]{0.31\linewidth}
\centering
\includegraphics[width=\linewidth]{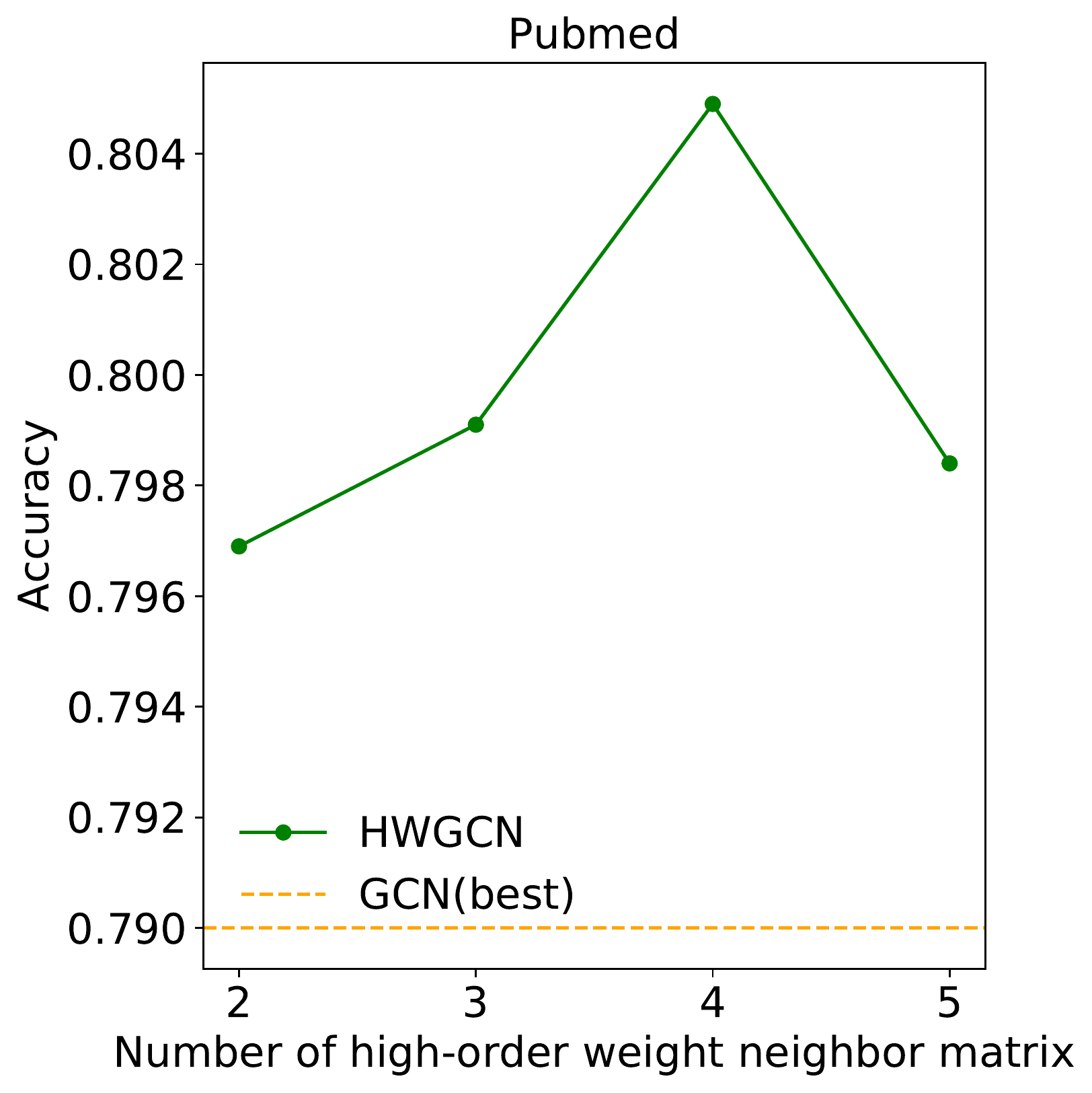}
\end{minipage}
}

\caption{Influence of neighbor matrix on classification performance (fixed split).}
\label{fixed 20}
\end{figure}

We also report the accuracy curve for training size $\in \{5, 10\}$ per class on the three datasets. The results are illustrated in Figure \ref{train 10} and Figure \ref{train 5} for another two random splits, and Figure \ref{fixed 20} for fixed split.

\newpage
\section{Results of matrix powers in replacement of distance matrix}
\begin{figure}[htbp]
\centering
\subfigure{
\begin{minipage}[t]{0.30\linewidth}
\centering
\includegraphics[width=\linewidth]{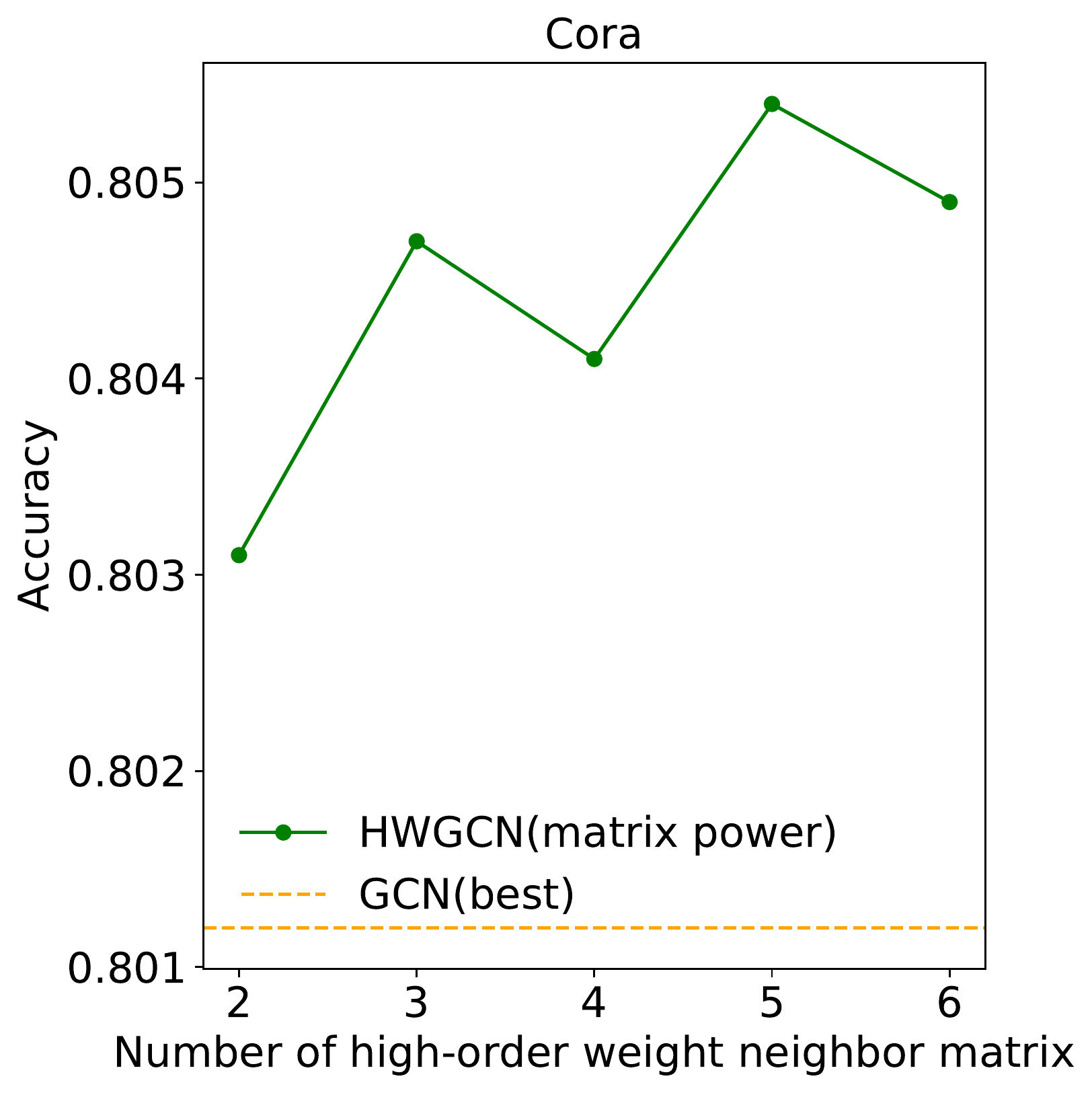}
\end{minipage}
}
\subfigure{
\begin{minipage}[t]{0.30\linewidth}
\centering
\includegraphics[width=\linewidth]{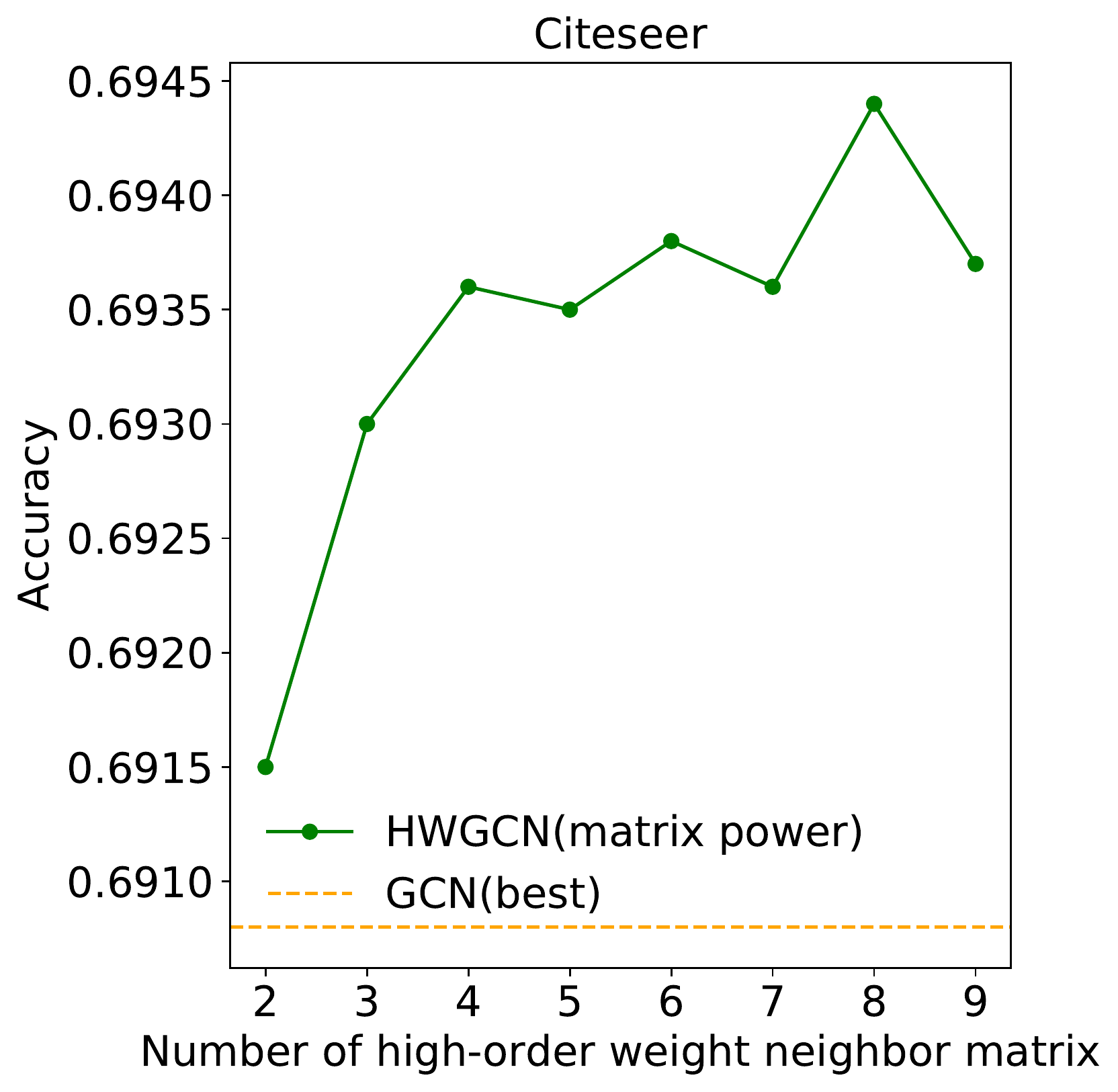}
\end{minipage}
}
\subfigure{
\begin{minipage}[t]{0.30\linewidth}
\centering
\includegraphics[width=\linewidth]{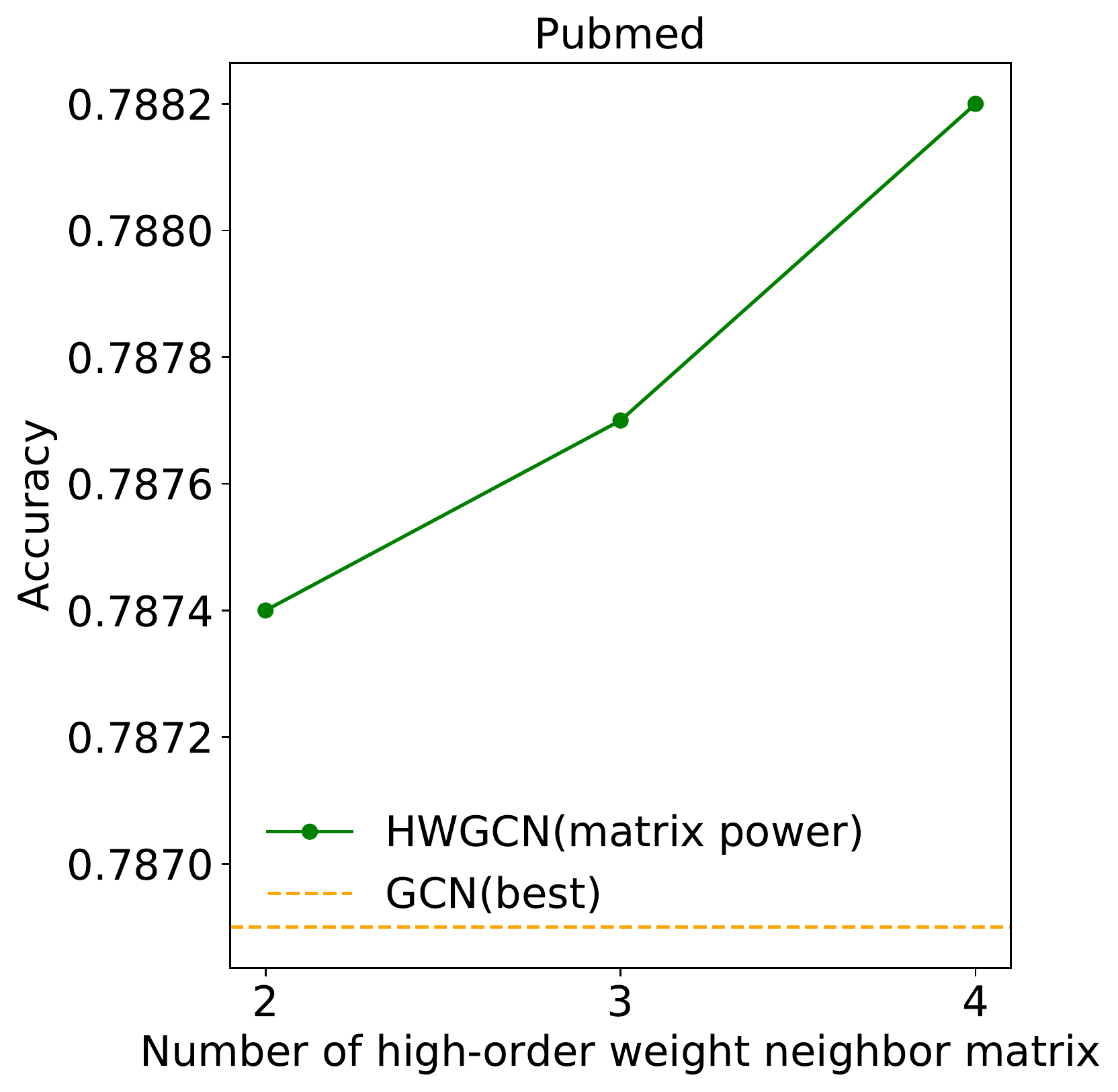}
\end{minipage}
}

\caption{Influence of matrix power on classification performance (random split with 20 per class).}
\label{power 20}
\vspace{-0.3cm}
\end{figure}

\begin{figure}[htbp]
\centering
\subfigure{
\begin{minipage}[t]{0.30\linewidth}
\centering
\includegraphics[width=\linewidth]{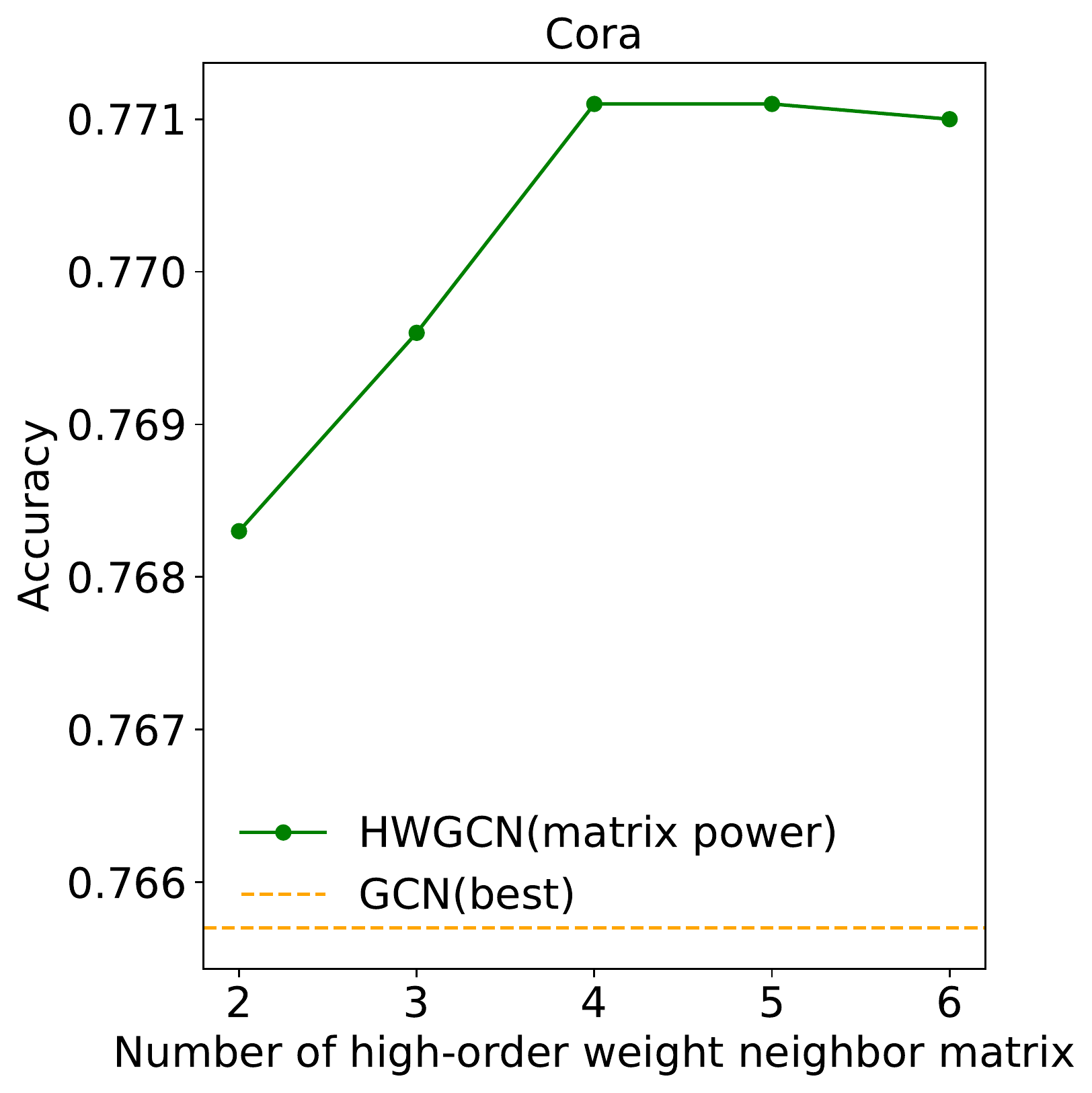}
\end{minipage}
}
\subfigure{
\begin{minipage}[t]{0.30\linewidth}
\centering
\includegraphics[width=\linewidth]{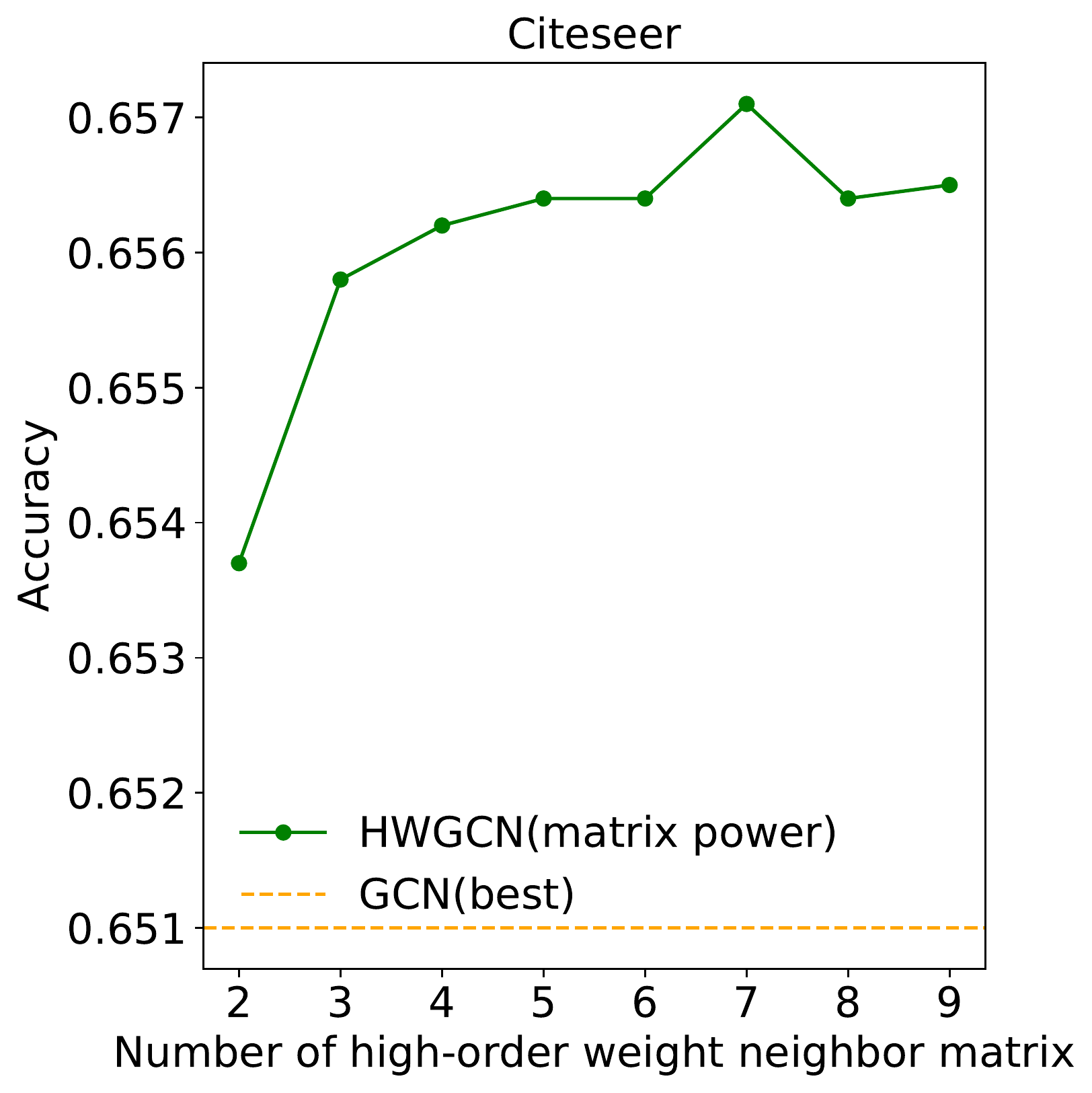}
\end{minipage}
}
\subfigure{
\begin{minipage}[t]{0.30\linewidth}
\centering
\includegraphics[width=\linewidth]{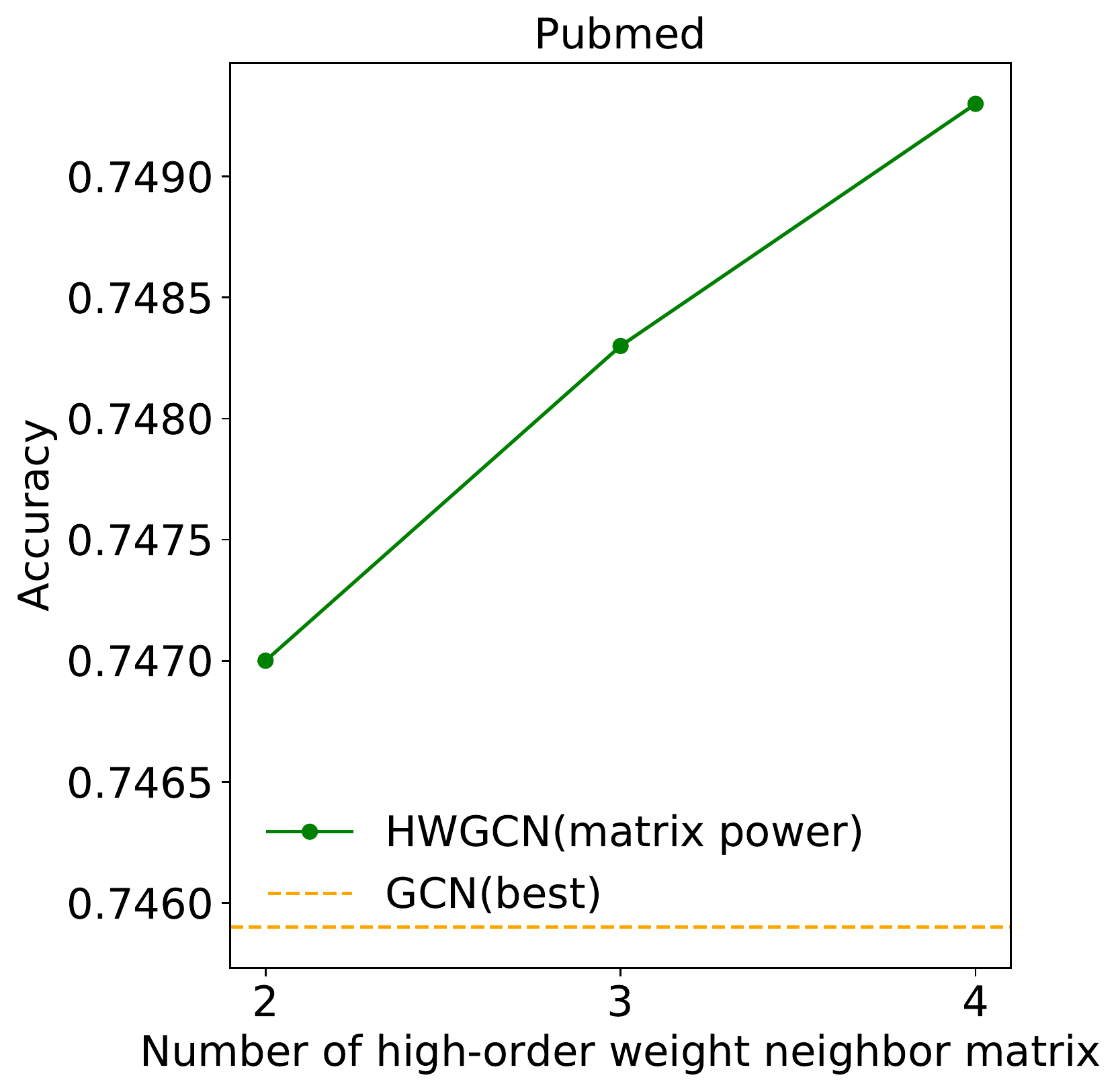}
\end{minipage}
}

\caption{Influence of matrix power on classification performance (random split with 10 per class).}
\label{power 10}
\vspace{-0.3cm}
\end{figure}

\begin{figure}[htbp]
\centering
\subfigure{
\begin{minipage}[t]{0.30\linewidth}
\centering
\includegraphics[width=\linewidth]{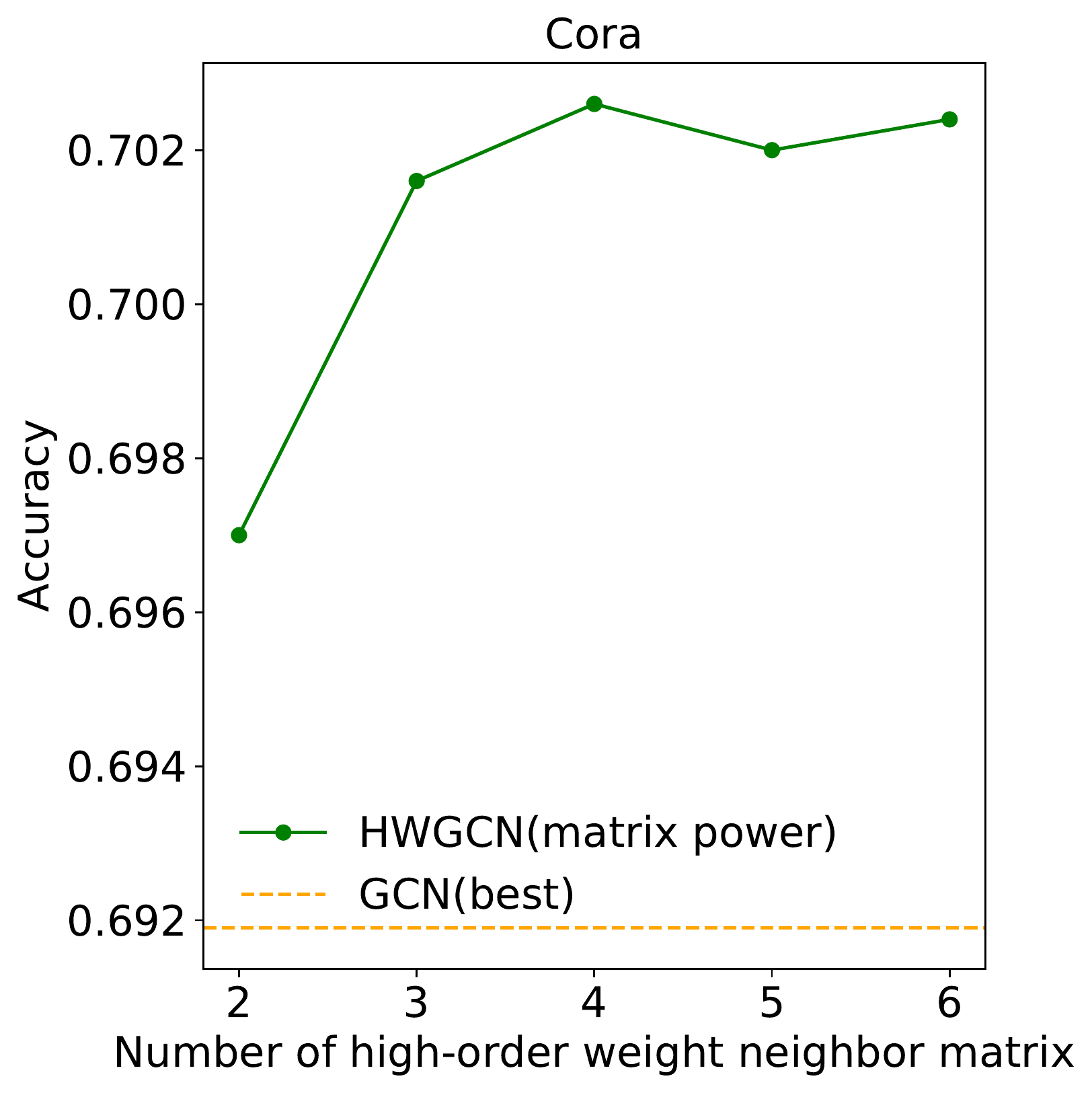}
\end{minipage}
}
\subfigure{
\begin{minipage}[t]{0.30\linewidth}
\centering
\includegraphics[width=\linewidth]{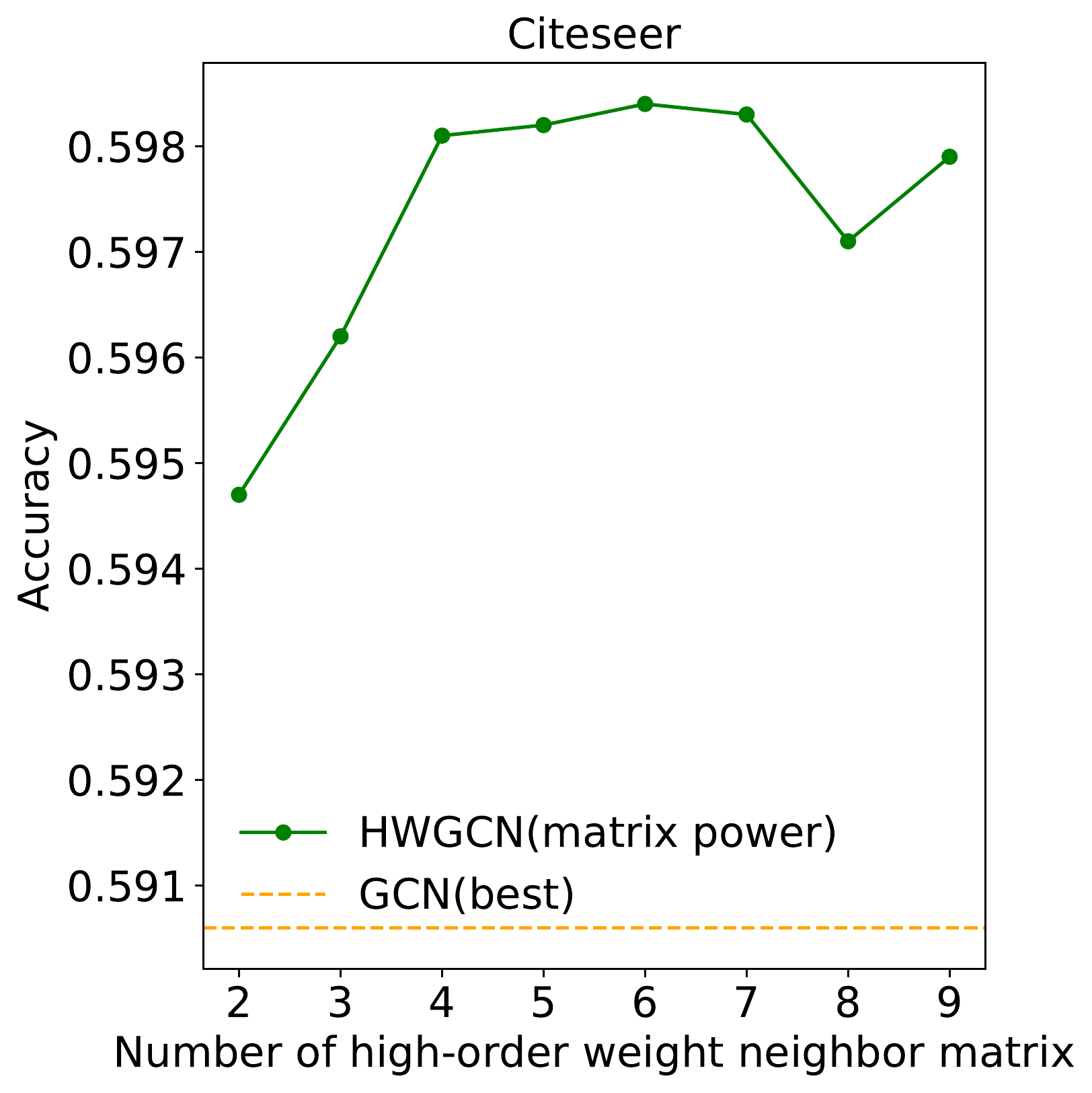}
\end{minipage}
}
\subfigure{
\begin{minipage}[t]{0.30\linewidth}
\centering
\includegraphics[width=\linewidth]{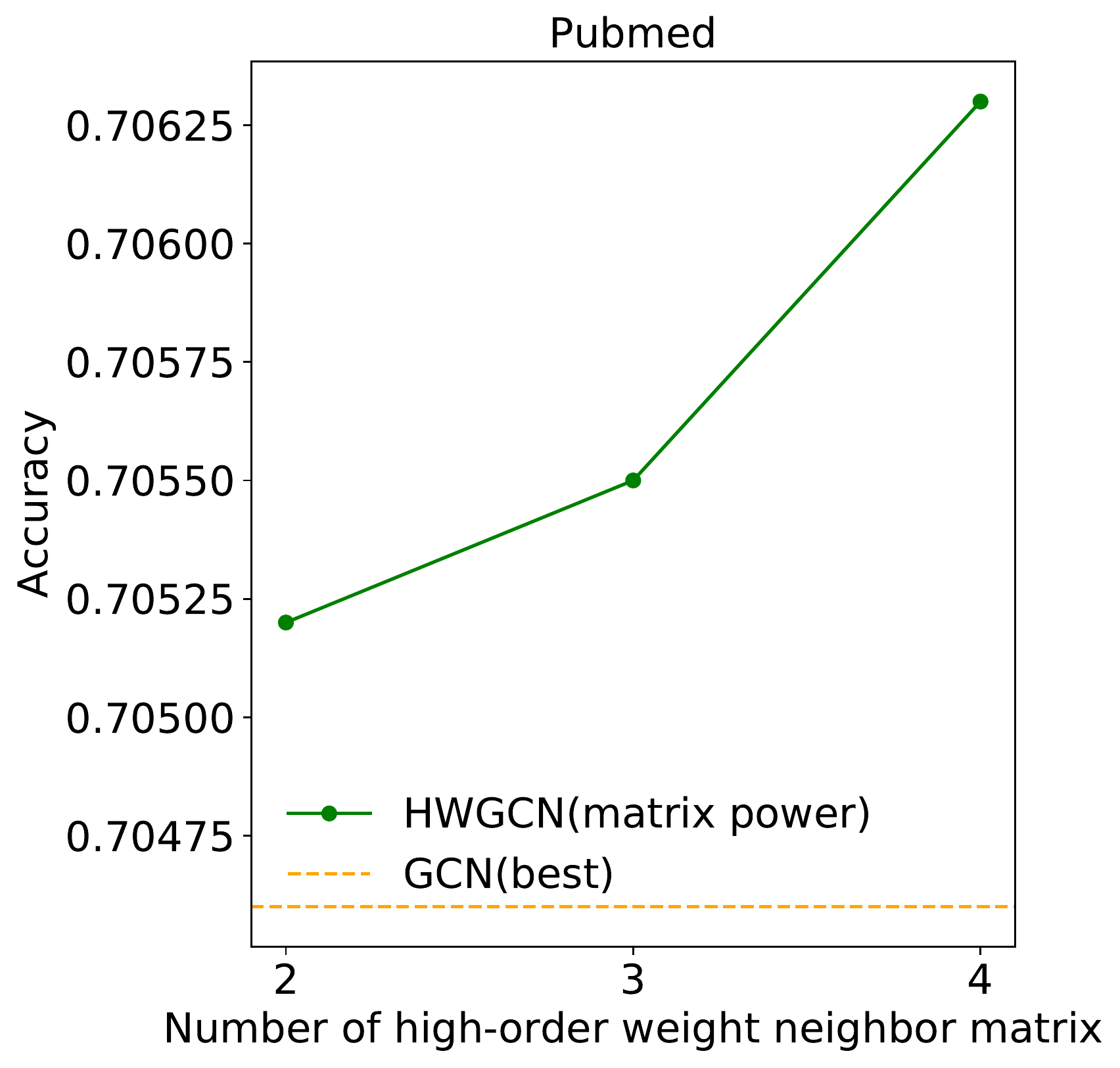}
\end{minipage}
}

\caption{Influence of matrix power on classification performance (random split with 5 per class).}
\label{power 5}
\vspace{-0.3cm}
\end{figure}

We replace the distance matrix with matrix powers and report the accuracy curve for training size $\in \{5, 10, 20\}$ per class on random splits of the three datasets. And the results are illustrated in Figure~\ref{power 20}, Figure~\ref{power 10} and Figure~\ref{power 5}.

\newpage
\section{The weight absolute value statistics} \label{sec:statistics}
Lasso with $l_1$-norm can get the sparse solution, and here we provide the weight absolute value statistics for $W^{(k)}$ in Table~\ref{cora sparsity statistics}, Table~\ref{citeseer sparsity statistics} and Table~\ref{pubmed sparsity statistics}. From the results, we can observe that large weight absolute values only take up a small portion.

\begin{table}[htbp]
\normalsize
\centering
\caption{The weight absolute value statistics of Cora  (\%)}
\setlength{\tabcolsep}{1.2mm}
\label{cora sparsity statistics}
\begin{tabular}{cccccccccccc}
\toprule
\textbf{$k^{th}$-order}&  $(0, 10^{-5})$     &  $(10^{-5}, 10^{-4})$  &   $(10^{-4}, 10^{-3})$  & $(10^{-3}, 10^{-2})$ & $(10^{-2}, 10^{-1})$ & $(10^{-1}, +\infty)$\\
\midrule
2   &42.79  &1.05  &1.91 & 16.37 & 35.09 & 2.79   \\
3   &63.21  &1.21  &1.83 & 13.64 & 19.51 & 0.60   \\
4   &79.39  &1.12  &1.46 & 9.04 & 8.80 & 0.19   \\
5   &85.57  &0.91  &1.19 & 6.81 & 5.44 & 0.08   \\
6   &86.74  &0.80  &1.16 & 6.44 & 4.80 & 0.06   \\
7   &84.73  &0.79  &1.26 & 7.37 & 5.78 & 0.07   \\
8   &79.47  &0.78  &1.57 & 9.66 & 8.42 & 0.10   \\
\bottomrule
\end{tabular}
\end{table}

\begin{table}[htbp]
\normalsize
\centering
\caption{The weight absolute value statistics of Citeseer  (\%)}
\setlength{\tabcolsep}{1.2mm}
\label{citeseer sparsity statistics}
\begin{tabular}{cccccccccccc}
\toprule
\textbf{$k^{th}$-order}&  $(0, 10^{-5})$     &  $(10^{-5}, 10^{-4})$  &   $(10^{-4}, 10^{-3})$  & $(10^{-3}, 10^{-2})$ & $(10^{-2}, 10^{-1})$ & $(10^{-1}, +\infty)$\\
\midrule
2   &24.07  &0.31  &1.51  & 15.92 & 50.80 & 7.39   \\
3   &43.06  &0.82  &2.08  & 18.30 & 34.40 & 1.34   \\
4   &55.59  &0.92  &2.14  & 17.43 & 23.52 & 0.40   \\
5   &62.46  &1.01  &2.09  & 16.16 & 18.08 & 0.20   \\
6   &66.37  &1.05  &2.07  & 15.11 & 15.28 & 0.12   \\
7   &70.16  &0.99  &2.00  & 14.02 & 12.75 & 0.08   \\
8   &73.59  &0.87  &2.03  & 13.18 & 10.28 & 0.05   \\
\bottomrule
\end{tabular}
\end{table}

\begin{table}[htbp]
\normalsize
\centering
\caption{The weight absolute value statistics of Pubmed  (\%)}
\setlength{\tabcolsep}{1.2mm}
\label{pubmed sparsity statistics}
\begin{tabular}{cccccccccccc}
\toprule
\textbf{$k^{th}$-order}&  $(0, 10^{-5})$     &  $(10^{-5}, 10^{-4})$  &   $(10^{-4}, 10^{-3})$  & $(10^{-3}, 10^{-2})$ & $(10^{-2}, 10^{-1})$ & $(10^{-1}, +\infty)$\\
\midrule
2   &55.29  &3.51  &1.36  & 11.81 & 25.54 & 2.49   \\
3   &85.23  &1.75  &1.04  & 5.83 & 5.78 & 0.37   \\
4   &94.04  &0.90  &0.69  & 2.69 & 1.61 & 0.07   \\
5   &95.80  &0.71  &0.57  & 1.93 & 0.96 & 0.03   \\
6   &96.41  &0.74  &0.57  &1.68  &0.58  &0.02    \\
7   &96.19  &0.74  &0.59  &1.80  &0.66  &0.02    \\
8   &95.61  &0.70  &0.65  &2.17  &0.85  &0.02    \\
\bottomrule
\end{tabular}
\end{table}

\end{document}

%% file: math_commands.tex
\usepackage{amsmath,amsfonts,bm}

\def\eqref#1{equation~\ref{#1}}

\def\1{\bm{1}}

\def\ervz{{\textnormal{z}}}

\DeclareMathAlphabet{\mathsfit}{\encodingdefault}{\sfdefault}{m}{sl}
\SetMathAlphabet{\mathsfit}{bold}{\encodingdefault}{\sfdefault}{bx}{n}

